%% file: iclr2026_conference.tex
\documentclass{article} 
\usepackage{iclr2026_conference,times}

\input{math_commands.tex}

\usepackage{hyperref}
\usepackage{url}
\usepackage{graphicx}
\usepackage{caption}
\usepackage{wrapfig}
\usepackage{amssymb}
\usepackage{algorithm}
\usepackage{algpseudocode}
\usepackage{setspace}
\usepackage{booktabs}
\usepackage{multirow}
\usepackage{amsthm}
\usepackage{pifont}

\usepackage{mathtools}

\newtheorem{property}{Property}
\newtheorem{definition}{Definition}
\hypersetup{colorlinks=true,linkcolor=red,citecolor=teal,urlcolor=cyan}

\title{Adaptive Debiasing Tsallis Entropy for Test-Time Adaptation}

\author{
Xiangyu Wu$^{1,2}$,~~Dongming Jiang$^{3}$,~~Feng Yu$^{1}$,~~Yueying Tian$^{4}$,~~Jiaqi Tang$^{5}$
\And
~~Qing-Guo Chen$^2$,~~Yang Yang$^1$\thanks{Corresponding author},~~Jianfeng Lu$^1$\myfootnote{Corresponding author} \\
$^1$Nanjing University of Science and Technology, $^2$Alibaba Cloud, $^3$University of Texas at Dallas \\
$^4$University of Sussex, $^5$Hong Kong University of Science and Technology \\
}

\author{\textbf{Xiangyu Wu}\thanks{Equal contribution.}~~$^{\spadesuit\diamondsuit}$\ \
        \textbf{Dongming Jiang}$^*$$^\clubsuit$\ \ 
        \textbf{Feng Yu}$^*$$^\spadesuit$\ \
        \textbf{Yueying Tian}$^\heartsuit$\ \ 
        \textbf{Jiaqi Tang}$^\triangle$ \\
        \textbf{Qing-Guo Chen}$^\diamondsuit$
        \textbf{Yang Yang}\thanks{Corresponding author.}~~$^\spadesuit$
        \textbf{Jianfeng Lu}$^\dagger$$^\spadesuit$ \\
  $^\spadesuit$Nanjing University of Science and Technology
  $^\clubsuit$University of Texas at Dallas \\
  $^\diamondsuit$Alibaba Cloud
  $^\heartsuit$University of Sussex 
  $^\triangle$Hong Kong University of Science and Technology
}

\newcommand{\myfootnote}[1]{\footnotemark[1]}

\iclrfinalcopy 
\begin{document}

\maketitle
\begin{abstract}
Mainstream Test-Time Adaptation~(TTA) methods for adapting vision-language models, \textit{e.g.}, CLIP, typically rely on Shannon Entropy~(\texttt{SE}) at test time to measure prediction uncertainty and inconsistency. However, since CLIP has a built-in bias from pretraining on highly imbalanced web-crawled data, \texttt{SE} inevitably results in producing biased estimates of uncertainty entropy. To address this issue, we notably find and demonstrate that Tsallis Entropy~(\texttt{TE}), a generalized form of \texttt{SE}, is naturally suited for characterizing biased distributions by introducing a non-extensive parameter $q$, with the performance of \texttt{SE} serving as a lower bound for \texttt{TE}. Building upon this, we generalize \texttt{TE} into \textbf{A}daptive \textbf{D}ebiasing \textbf{T}sallis \textbf{E}ntropy~(\texttt{ADTE}) for TTA, customizing a class-specific parameter $q^l$ derived by normalizing the estimated label bias from continuously incoming test instances, for each category. This adaptive approach allows \texttt{ADTE} to accurately select high-confidence views and seamlessly integrate with label adjustment strategy to enhance adaptation, without introducing distribution-specific hyperparameter tuning. Besides, our investigation reveals that both \texttt{TE} and \texttt{ADTE} can serve as direct, advanced alternatives to \texttt{SE} in TTA, without any other modifications. Experimental results show that \texttt{ADTE} outperforms state-of-the-art methods on ImageNet and its five variants, and achieves the highest average performance on $10$ cross-domain benchmarks, regardless of the model architecture or text prompts used. Our code is available at \url{https://github.com/Jinx630/ADTE}.
\end{abstract}

\section{Introduction}
Vision-Language Models~(VLMs)~\citep{VLMs-Openai-CLIP,q2v,VLMs-XVLM2,TaAM-CPT,RPRCR,RML,yu2025visual,zeng2025janusvln}, pretrained on large-scale datasets~\citep{ITP-CC3M,ITP-Laion-5b}, exhibit remarkable generalization abilities across various downstream tasks. Despite this, they are susceptible to performance degradation when confronted with considerable discrepancies between training and testing domains. To mitigate this, a technology known as Test-Time Adaptation~(TTA)~\citep{TTA-TPT,TTA-Frolic,TTA-BCA,TTA-ML-TTA}, enables models to adapt instantaneously to diverse instance distributions during testing, in contrast to earlier prompt learning techniques~\citep{Prompt-CoOp,Prompt-DualCoOp++,Prompt-VLPL,PVP,Refining} that require complex training procedures.

Among these representative TTA works, TPT~\citep{TTA-TPT} and its enhancement, DiffTPT~\citep{TTA-DiffTPT}, learn instance-level prompts by selecting high-confidence augmented views for each test instance. Zero~\citep{TTA-Zero} simplifies the TTA process, demonstrating that the prediction of the marginal probability distribution remains approximately invariant under entropy minimization. ML-TTA~\citep{TTA-ML-TTA}, equipped with Bound Entropy Minimization~(\texttt{BEM}), enables the adaptation of multi-label instances. The central idea is to select lower entropy views as the high-confidence views, aiming to reduce uncertainty among these views, a readily demonstrable theory.

\begin{figure*}
 \centering
 \includegraphics[scale=0.42]{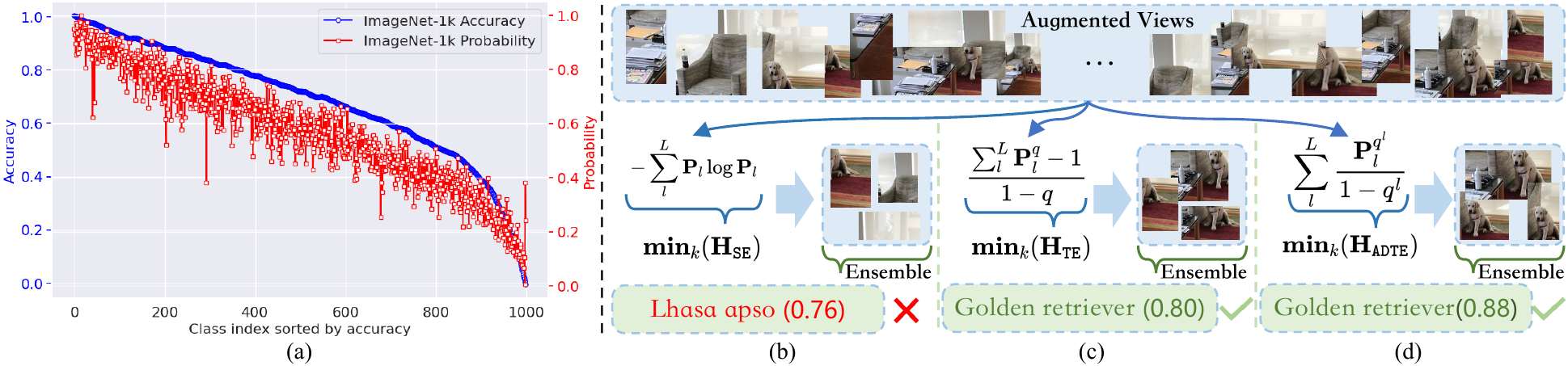}
 \captionsetup{font={footnotesize}}
 \caption{(a) VLM bias, showing higher confidence and accuracy for \texttt{head} classes and lower confidence and accuracy for \texttt{tail} classes. (b) The standard Shannon Entropy~(\texttt{SE})-based method is widely used in TTA. (c) and (d) Our proposed method, which uses Tsallis Entropy~(\texttt{TE}) and Adaptive Debiasing Tsallis Entropy~(\texttt{ADTE}) for selecting high-confidence views.}
 \label{fig:method-comparsion}
 \vspace{-1.5em}
\end{figure*}


However, VLMs~(\textit{e.g.}, CLIP), as discussed in research~\citep{bias-01,bias-02,bias-03,TTA-Frolic,wu2025evaluation,ke2025early}, are pretrained on imbalanced web-scale datasets, inevitably possessing inherent prediction bias. This bias causes the model to consistently exhibit \texttt{low/high} confidence in the \texttt{tail/head} categories, leading to \texttt{lower/higher} accuracy, as shown in Figure~\ref{fig:method-comparsion}~(a). For this reason, during the TTA process, the predicted probability for a category $l$, denoted as $\mathbf{p}$, may significantly differ from its true unbiased probability $\mathbf{\hat{p}}$~(\textit{i.e.}, $\vert \mathbf{p}-\mathbf{\hat{p}}\vert>0$). Methods based on Shannon Entropy~(\texttt{SE}), defined as $\mathbf{H}_{\texttt{SE}}=-\sum_l \mathbf{p}\log \mathbf{p}$, are impacted because their entropy calculations rely on these potentially biased probabilities $\mathbf{p}$. Moreover, from its definition, \texttt{SE} fails to account for varying degrees of bias in probabilities across different classes (\textit{i.e.,} \texttt{head}, \texttt{middle}, and \texttt{tail} classes). Instead, \texttt{SE} applies a uniform computation formula~($-\mathbf{p}\log \mathbf{p}$) across all probabilities, as shown in Figure~\ref{fig:method-comparsion}~(b). As a result, in TTA methods that rely on \texttt{SE}, the entropy values estimated for each augmented view are themselves biased. This bias, in turn, affects the selection of high-confidence views from a set of augmented views.

Our first key observation is that Tsallis Entropy~(\texttt{TE})~\citep{Tsallis-01,Tsallis-02}, a generalization of \texttt{SE}, is well-suited for characterizing uncertainty in the presence of biased distributions. By introducing an additional parameter $q$\footnote{$q$ is called \textit{non-extensive parameter} in the literature~\citep{Tsallis-01,Tsallis-02}.}, \texttt{TE} can characterize likelihoods exhibiting statistical dependence and effectively mitigate the impact of bias. First, we demonstrate that \texttt{TE} is a limiting case of \texttt{SE} when $q \to 1$. On the other hand, when $q < 1$, \texttt{TE} tends to select more confident views (characterized by higher $\mathbf{Tcr}_K$ in Definition~\ref{tcr_k}), implying that the performance of \texttt{SE} can be regarded as a lower bound for that of \texttt{TE}. In other words, there must exist an appropriate parameter $\hat{q}$ such that models using \texttt{TE} outperform those using \texttt{SE} in the selection of high-confidence views. Subsequently, we demonstrate that when $q<1$, \texttt{TE} can effectively alleviate the impact of inherent bias for VLMs.

However, there are two major difficulties in applying \texttt{TE} in TTA practice: \textbf{(i)} manually tuning the optimal value of $q$ in standard \texttt{TE} is impractical for various test distributions; and \textbf{(ii)} the optimal parameter $\hat{q}$ for each category may also differ, since each category is affected by bias differently, as discussed earlier. Therefore, we generalize \texttt{TE} into \textbf{A}daptive \textbf{D}ebiasing \textbf{T}sallis \textbf{E}ntropy~(\texttt{ADTE}) for TTA, customizing a class-specific $q^l$ for category $l$, as shown in Figure~\ref{fig:method-comparsion}~(d). These parameters are derived via min–max normalization of the estimated label biases from continuously incoming test instances. Significantly, both \texttt{TE} and \texttt{ADTE} can serve as direct, advanced alternatives to \texttt{SE} and integrate seamlessly with the logit adjustment strategy to enhance adaptation performance. Experiments show that, irrespective of the model architecture or text prompts employed, \texttt{ADTE} surpasses the SOTAs on ImageNet and its five variants, achieving the highest average performance across $10$ cross-domain benchmarks.

\section{Related Works}

\textbf{Test-Time Adaptation~(TTA)}. TTA methods~\citep{TTA-TPT,TTA-TDA,TTA-Frolic,TTA-ML-TTA,TTA-BCA} dynamically adjust pre-trained models using unlabeled test data during inference, \textit{e.g.}, detection~\citep{ouyang2024learn,zhao2024balf,edstedt2024dedode,zhao2026non} tasks. TTA has been explored in various settings, including ``fully'' TTA~\citep{TTA-Tent,TTA-Delta}, ``online'' TTA~\citep{TTA-Stationary,TTA-NotEnough}, and ``continuous'' TTA~\citep{TTA-VIDA,TTA-EcoTTA}. TPT~\citep{TTA-TPT} is one of the first works to apply prompt tuning for adapting VLMs to previously unseen distributions. Zero~\citep{TTA-Zero} demonstrates that minimizing entropy does not alter the dominant class prediction, providing a theoretical basis for this approach. Other works have introduced new techniques: Frolic~\citep{TTA-Frolic} uses label-free prompt distribution learning and bias correction, ML-TTA~\citep{TTA-ML-TTA} employs a bound entropy minimization objective in multi-label scenarios, and BCA~\citep{TTA-BCA} leverages Bayesian principles to refine predictions as new data arrives.

\textbf{Entropy-based Uncertainty Minimization in Adaptation.}
Entropy Minimization (EM)~\citep{EM-01,EM-02,EM-04,Facilitating,CoVLR,CMDGK,SePer,zeng2025FSDrive} is a common strategy for reducing prediction uncertainty and promoting clearer decision boundaries. Early work~\citep{EM-01} used minimum entropy regularization to leverage unlabeled data, while MME~\citep{EM-03} and Tent~\citep{TTA-Tent} showed its effectiveness for adapting to varied test distributions. Recently, Tsallis Entropy (TE) has emerged as an alternative with a tunable parameter $q$ controlling the sharpness of the entropy landscape. In domain adaptation and self-training~\citep{lu2023meta,liu2021cycle,leeduet,zhao2023source}, TE has been optimized to improve pseudo-label reliability and robustness to noise in supervised or source-free settings, focusing on feature representation learning. While EM minimizes standard Shannon entropy, TE offers a more flexible formulation that may extend to online test-time adaptation.

\textbf{Logit Adjustment~(LA)}. LA~\citep{LA-01,LA-02,LA-03,LA-04,LA-05,LA-06} is primarily used for long-tailed recognition and class-imbalanced learning. It adjusts a model's logits to compensate for biases caused by imbalanced training data. GCL~\citep{LA-02} adds a Gaussian perturbation to logits, giving larger perturbations to \texttt{tail} classes to enhance their gradient contribution. LoTNext~\citep{LA-04} introduces graph-based and long-tailed loss adjustments to improve spatial and temporal prediction. HTC~\citep{LA-05} addresses the long-tail issue with candidate label set disambiguation, class distribution estimation, and classifier weight estimation. COCL~\citep{LA-06} combines debiased large-margin learning and outlier-class-aware logit calibration to effectively mitigate biases. We employ the logit adjustment to estimate the class-specific parameter $q^l$ for each category in Tsallis Entropy.

\section{Preliminaries}
\label{entropy minimization}
Consider the CLIP~\citep{VLMs-Openai-CLIP} model pretrained on $\mathcal{D}^{\text{train}} \coloneqq \{ (\mathbf{x}_i^{\text{train}},\mathbf{y}_i^{\text{train}})\mid\mathbf{x}_i^{\text{train}}\!\in\!\mathcal{X}^{\text{train}}, \mathbf{y}_i^{\text{train}}\!\in\!\mathcal{Y}^{\text{train}}\}_{i=1}^{M^{\text{train}}}$ and a downstream test set $\mathcal{D}^{\text{test}} \coloneqq \{ (\mathbf{x}_i^{\text{test}},\mathbf{y}_i^{\text{test}})\mid\mathbf{x}_i^{\text{test}}\!\in\!\mathcal{X}^{\text{test}}, \mathbf{y}_i^{\text{test}}\!\in\!\mathcal{Y}^{\text{test}}\}_{i=1}^{M^{\text{test}}}$, which may follow an arbitrary distribution. We illustrate the standard Test-Time Adaptation~(TTA) process with Zero~\citep{TTA-Zero}, which consists of \textit{Random view augmentation}, \textit{Confident views selection}, and \textit{Confident views ensemble}.

\textbf{\textit{Random View Augmentation}}. Given a test instance $\mathbf{x}^{\text{test}}$ from $\mathcal{D}^{\text{test}}$ and a set $\mathcal{A}$ of $N$ random augmentation functions, $\mathbf{x}^{\text{test}}$ is augmented $N$ times to create a set of diverse views, which are denoted as $\mathbf{X}^{\text{test}} \coloneqq \{ \mathbf{x}_j^{\text{test}}\mid\mathbf{x}_j^{\text{test}}=\mathcal A_j(\mathbf{x^{\text{test}}})\}_{j=1}^{N}$.

\textbf{\textit{Confident Views Selection}}. In information theory, Shannon Entropy~(\texttt{SE},~\citep{Shannon-Entropy}) is commonly used to quantify uncertainty\footnote{Thermodynamic entropy, an earlier concept, is typically expressed via Boltzmann-Gibbs-Shannon entropy $\mathbf{H}_\texttt{BGS} = -k \sum_i p_i \log p_i$, where $k$ is the Boltzmann constant. $\mathbf{H}_\texttt{BGS}$ is mathematically identical to Shannon Entropy, but generally less familiar to the machine learning community}. For each view, uncertainty is measured by \texttt{SE}, defined as:
\begin{equation}
    \mathbf{H}_\texttt{SE}(\mathbf{P}(\cdot\mid\mathbf{x}_j^{\text{test}})) = -\sum_{l=1}^{L} \mathbf{P}(y = l\mid\mathbf{x}_j^{\text{test}}) \log[\mathbf{P}(y = l\mid\mathbf{x}_j^{\text{test}})],
\end{equation}
where $l \in \mathcal{Y}^{\text{test}}$ denotes a class label, and $L$ is the number of classes in the test set. The core of Zero~\citep{TTA-Zero} lies in minimizing the marginal entropy over the prediction distributions corresponding to the selected high-confidence augmented views~(\textit{i.e.,} views with lower entropy) by a ratio $\tau$, which reduces the model's uncertainty and prediction inconsistency across these views. However, due to the inherent prediction bias of models like CLIP, the estimated \texttt{SE} values for \texttt{head} or \texttt{tail} categories are themselves biased. This bias in \texttt{SE} estimation, in turn, affects the selection of high-confidence views.

\textbf{\textit{Confident Views Ensemble}}. Unlike TPT~\citep{TTA-TPT}, which uses high-confidence views to update the model's prompts during TTA, Zero~\citep{TTA-Zero} simplifies the TTA process. It argues that the updated model's prediction remains invariant under entropy minimization, which allows the final prediction to be obtained by directly aggregating the high-confidence views. The model then immediately adapts to the next test instance. Due to its simplicity and effectiveness, \texttt{SE} has become a standard metric for high-confidence view selection in recent TTA methods.

\section{Method}
\subsection{Tsallis Entropy}
\label{Tsallis Entropy}
When multiple probability distributions are involved, Shannon entropy $\mathbf{H}_\texttt{SE}$ quantifies total uncertainty under the~\textit{extensivity} assumption, meaning that the entropy of independent parts simply adds, i.e., $\mathbf{H}_\texttt{SE}(\{A, B\}) = \mathbf{H}_\texttt{SE}(A) + \mathbf{H}_\texttt{SE}(B)$. In the real world, such as with biased model predictions, this additive property does not hold. To capture such \textit{non-extensivity}, \texttt{TE} generalizes \texttt{SE} by introducing a parameter $q$ that characterizes the degree of non-additivity, defined as:
\begin{equation}\label{eq:Tsallis}
    \mathbf{H}_\texttt{TE}(\mathbf{P}(\cdot\mid \mathbf{x}_j^{\text{test}})) = \frac{1}{1-q}\left( \sum_{l=1}^L \mathbf{P}(y=l\mid \mathbf{x}_j^{\text{test}})^q - 1 \right),
\end{equation}
where $q \in \mathbb{R}$ is the additional hyperparameter. Using \texttt{TE}, the total entropy can be calculated as $\mathbf{H}_\texttt{TE}(\{A, B\}) = \mathbf{H}_\texttt{TE}(A) + \mathbf{H}_\texttt{TE}(B) + (1 - q) \mathbf{H}_\texttt{TE}(A) \mathbf{H}_\texttt{TE}(B)$, where the additional term $(1 - q) \mathbf{H}_\texttt{TE}(A) \mathbf{H}_\texttt{TE}(B)$ is used to characterize influence between components. We first demonstrate that \texttt{TE} possesses the following crucial properties, which make \texttt{SE} the lower bound in performance for \texttt{TE}. In other words, there exists an appropriate parameter $\hat{q}$ for which models based on \texttt{TE} are capable of selecting more accurate and confident views compared to those relying on \texttt{SE}.

$\blacktriangleright$ \texttt{Top-K Cumulative Reliability:}
\begin{definition}\label{tcr_k}
    VLMs like CLIP classify images by computing similarity scores between an image and ``a photo of $\{\texttt{classes}\}$''. For an augmented view $\mathbf{x}_j^{\textup{test}}$, the similarity with a class $l$ is $\mathbf{z}_l^{\top}\mathbf{x}_j^{\textup{test}}$, where $\mathbf{z}_l^{\top}$ denotes the textual embedding representing class $l$. We denote the Top-K Cumulative Reliability $\mathbf{Tcr}_K$ as the sum of the $K$ highest similarity scores. Mathematically, it is given by:
    \begin{equation}
        \mathbf{Tcr}_K(\mathbf{x}_j^{\textup{test}}) = \sum_{l \in \mathcal{T}_K} \mathbf{z}_l^{\top}\mathbf{x}_j^{\textup{test}},
    \end{equation}
    where $\mathcal{T}_K$ is the set of indices corresponding to the $K$ highest similarity scores. Actually, CLIP typically achieves an extremely high Top-K~(e.g., K=5 or K=10) accuracy, indicating that although CLIP may not always predict the exact label, it is proficient at generating a candidate set that contains the correct label.
\end{definition}

\begin{wrapfigure}{r}{0.39\textwidth}
    \centering
    \vspace{-1.7em}
    \includegraphics[width=0.39\textwidth]{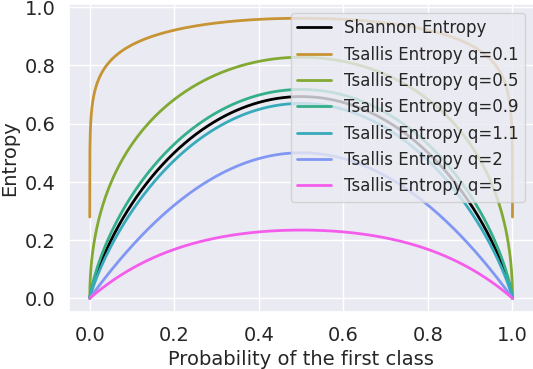}
    \vspace{-2em}
    \captionsetup{font={footnotesize}}
    \caption{Comparison between \texttt{SE} and \texttt{TE}.}
    \vspace{-1em}
    \label{fig:limits}
\end{wrapfigure}
$\blacktriangleright$ \texttt{Shannon-Tsallis $q \to 1$ Equivalence:}
\begin{property}\label{property-1}
As $q \to 1$, \texttt{TE} becomes equivalent to \texttt{SE}:
\begin{equation}
    \lim_{q \to 1}\mathbf{H}_{\textup{\texttt{TE}}}(\mathbf{P}(\cdot\mid\mathbf{x}_j^{\textup{test}})) = \mathbf{H}_\textup{\texttt{SE}}(\mathbf{P}(\cdot\mid\mathbf{x}_j^{\textup{test}})).
\end{equation}
\textit{See the Appendix~\ref{Proof of property 1} for the detailed proof.} \textup{Property~\ref{property-1} demonstrates that \texttt{TE} is a generalization of \texttt{SE}, as illustrated in Figure~\ref{fig:limits} for a case with two classes. This property confirms that \texttt{TE} is consistent with the well-established \texttt{SE} theory and can be used to analyze more intricate non-extensive probability distributions by adjusting $q$.}
\end{property}

$\blacktriangleright$ \texttt{Higher $\mathbf{Tcr}_K$ under TE as $q\searrow$ and Comparison with SE:}
\begin{property}\label{property-2}
Through experimental analysis, we find that as the parameter $q$ decreases, the set of high-confidence views selected by \texttt{TE} tends to have a higher average $\mathbf{Tcr}_K$ value (for $K > 1$). For two different parameter values $q_1$ and $q_2$ with $q_1 < q_2$, and their corresponding selected view sets $\mathcal{X}_{q_1}$ and  $\mathcal{X}_{q_2}$ of equal size, we observe:
\begin{equation}
    \frac{1}{|\mathcal{X}_{q_1}|}\sum_{\mathbf{x_1}\in \mathcal{X}_{q_1}} \mathbf{Tcr}_{K}(\mathbf{x_1}) > \frac{1}{|\mathcal{X}_{q_2}|} \sum_{\mathbf{x_2}\in \mathcal{X}_{q_2}} \mathbf{Tcr}_{K}(\mathbf{x_2}).
\end{equation}
Furthermore, there exists a particular $q^*$ and corresponding view set $\mathcal{X}_{q^*}^{\texttt{\textup{TE}}}$, \texttt{TE} outperforms \texttt{SE} in this regard, with the view set selected by \texttt{SE} denoted as $\mathcal{X}^{\texttt{\textup{SE}}}$:
\begin{equation}
     \frac{1}{|\mathcal{X}_{q^*}^{\texttt{\textup{TE}}}|} \sum_{\mathbf{x_1}\in \mathcal{X}_{q^*}^{\texttt{\textup{TE}}}} \mathbf{Tcr}_{K}(\mathbf{x_1}) > \frac{1}{|\mathcal{X}^{\texttt{\textup{SE}}}|} \sum_{\mathbf{x_2}\in \mathcal{X}^{\texttt{\textup{SE}}}} \mathbf{Tcr}_{K}(\mathbf{x_2}).
\end{equation}
\textit{See the Appendix~\ref{proof of property 2} for the detailed experimental analysis.} \textup{Property~\ref{property-2} indicates that, in contrast to \texttt{SE}, \texttt{TE} tends to select views with higher $\mathbf{Tcr}_K$, and the corresponding ground truth has a higher similarity level. The core objective of TTA is to select the most confident and accurate views.}
\end{property}

\subsection{Correcting Biased Entropy Effect with \texttt{TE}}\label{Correcting Biased Entropy}
With the theoretical properties of \texttt{TE} established, we now investigate how its additional parameter $q$ can mitigate the biased entropy effect observed with \texttt{SE} in VLMs. For a \texttt{tail} category, the biased prediction $\mathbf{p}$ and unbiased prediction $\mathbf{\hat{p}}$ are both close to $0$, with $\mathbf{\hat{p}} > \mathbf{p}$ (based on empirical results; this is likely due to the CLIP model being very uncertain about \texttt{tail} predictions, as shown in Figure~\ref{fig:method-comparsion}~(a)). This results in a biased entropy value, where $(-\mathbf{\hat{p}}\log \mathbf{\hat{p}})-(-\mathbf{p}\log \mathbf{p})>0$.

To analyze the effect of \texttt{TE}, we rewrite Equation~\ref{eq:Tsallis} to examine the entropy value of each category:
\begin{equation}
\begin{aligned}
    \mathbf{H}_\texttt{TE}(\mathrm{P}) &= \frac{\sum_{l = 1}^{L}\mathrm{P}_l^q - 1}{1 - q} = \sum_{l = 1}^{L} \frac{\mathrm{P}_{l}^{q}}{1 - q} - \frac{1}{1-q} =\sum_{l = 1}^{L} \frac{\mathrm{P}_{l}^{q}}{1 - q} - \mathbf{C},
\end{aligned}
\label{eq:TE value}
\end{equation}
where $\mathrm{P}_l^q \coloneqq \mathbf{P}(y=l\mid\mathbf{x}_j^{\text{test}})^q$, and $\mathbf{C}$ is a constant. Since the constant $\mathbf{C}$ does not affect the ranking of entropy values, we can focus only on the term $\frac{\mathbf{p}^q}{1-q}$. For \texttt{TE} to correct the bias of \texttt{SE}, we require the inequality $\underline{\mathbf{F}(\mathbf{p},q)=\frac{\mathbf{p}^q}{1-q}-(-\mathbf{p}\log \mathbf{p})>0}$ to hold. The numerical value of $\mathbf{F}(\mathbf{p},q)$ reflects the degree to which \texttt{TE} corrects the bias entropy of \texttt{SE}. We analyze the behavior of this function for a \texttt{tail} category probability $\mathbf{p} \in (0, \epsilon)$, where $\epsilon > 0$ is a small value:

{\footnotesize
    $\blacktriangleright$ \texttt{(1).As $q\!\!\to\!\!+\infty,$ we have $\lim_{\mathbf{p}\to0^+}\mathbf{F}(\mathbf{p},q)=0^-$ and $\forall \!q_1\!<\!q_2\!\!<\!+\infty, \Rightarrow\!\mathbf{F}(\mathbf{p}, q_1)\!\!<\!\!\mathbf{F}(\mathbf{p}, q_2)\!\!<\!\!0$.}
    
    $\blacktriangleright$ \texttt{(2).As $q\!\to\!1^{+},$ we have $\lim_{\mathbf{p}\to0^+}\mathbf{F}(\mathbf{p},q)\!=\!-\infty$ and $\forall 1\!<\!q_1\!<\!q_2, \Rightarrow\!\mathbf{F}(\mathbf{p}, q_1)\!<\!\mathbf{F}(\mathbf{p}, q_2)\!<\!0$.}
    
    $\blacktriangleright$ \texttt{(3).As $q\!\to\!0^{+},$ we have $\lim_{\mathbf{p}\to0^{+}}\mathbf{F}(\mathbf{p},q)\!=\!1^-$ and $\forall 0\!<\!q_1\!<\!q_2, \Rightarrow\!\mathbf{F}(\mathbf{p}, q_1)\!>\!\mathbf{F}(\mathbf{p}, q_2)\!>\!0$.}
    
    $\blacktriangleright$ \texttt{(4).As $q\!\to\!\!1^-,$ we have $\lim_{\mathbf{p}\to0^+}\mathbf{F}(\mathbf{p},q)\!=\!+\infty$ and $\forall q_1\!<\!q_2\!<1, \Rightarrow\!\mathbf{F}(\mathbf{p}, q_2)\!>\!\mathbf{F}(\mathbf{p}, q_1)\!>\!0$.}
}

\textit{See the Appendix~\ref{proof of Correcting Biased Entropy} for the detailed proof}. Based on Property~\ref{property-1}~(when $q\to1$, \texttt{TE} and \texttt{SE} become equivalent), conclusions $(2)$ and $(4)$ are not applicable. Conclusion $(1)$ also does not meet our requirements since $\mathbf{F}(\mathbf{p},q)<0$. Conclusion $(3)$ shows that when $0 < q < 1$, \texttt{TE} can naturally mitigate the effect of VLM bias, with the correction magnitude increasing as $q$ decreases.

\begin{figure}[h]
    \centering
    \includegraphics[width=1\textwidth]{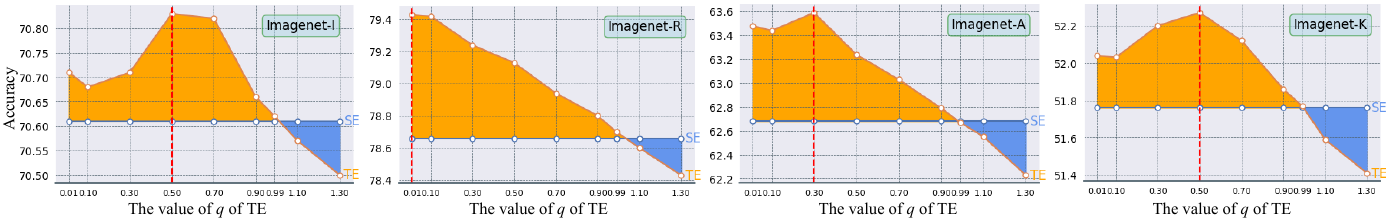}
    \vspace{-1em}
    \captionsetup{font={footnotesize}}
    \caption{\texttt{TE} at different $q$ values vs. \texttt{SE}; the red dashed line marks the optimal $q$ of \texttt{TE}.}
    \label{fig:Performance comparison of TE and SE}
\end{figure}

\subsection{Adaptive Debiasing Tsallis Entropy~(ADTE)}
A limitation of the standard \texttt{TE} formulation is that $q$ is a manually-tuned hyperparameter. A value of $q > 1$ can exacerbate the bias, while a value too close to $0$ can overcorrect it. As shown in Figure~\ref{fig:Performance comparison of TE and SE}, the optimal $q$ varies across different test distributions, making manual tuning impractical in a streaming environment. To address this, we generalize \texttt{TE} into \textbf{A}daptive \textbf{D}ebiasing \textbf{T}sallis \textbf{E}ntropy~(\texttt{ADTE}), which customizes a class-specific parameter $q^l$ for each category $l$, allowing the model to adapt to any test distribution. Building upon Equation~\ref{eq:TE value}, \texttt{ADTE} is defined as: 
\begin{equation}
    \mathbf{H}_\texttt{ADTE}(\mathrm{P}) = \sum_{l=1}^{L} \frac{\mathrm{P}_{l}^{q^l}}{1-q^l}.
    \label{equ:ADTE}
\end{equation}
Based on our analysis in Section~\ref{Correcting Biased Entropy}, when $0 < q < 1$, \texttt{TE} can alleviate the impact of bias, and a smaller value of $q^l$ results in a greater degree of correction. The magnitude of this correction depends directly on the prediction bias of the category. Therefore, we can deduce the relationship between the prediction bias and the corresponding $q^l$ value for each category:
\begin{footnotesize}
\begin{equation}
\left.
\begin{array}{l}
\blacktriangleright \texttt{Head class }l^\mathrm{H}, \text{dis}[\mathbf{P}_{l^\mathrm{H}},\mathbf{\hat{P}}_{l^\mathrm{H}}]>0, \mathbf{P}_{l^\mathrm{H}} \to 1,\mathbf{P}_{l^\mathrm{H}}>\mathbf{\hat{P}}_{l^\mathrm{H}} \\
\blacktriangleright \texttt{Tail class }l^\mathrm{T}, \text{dis}[\mathbf{P}_{l^\mathrm{T}},\mathbf{\hat{P}}_{l^\mathrm{T}}]>0, \mathbf{P}_{l^\mathrm{T}} \to 0,\mathbf{P}_{l^\mathrm{T}}<\mathbf{\hat{P}}_{l^\mathrm{T}} \\
\end{array}
\right\}
\begin{matrix}
\Rightarrow -\mathbf{p}\log(\mathbf{p})<-\mathbf{\hat{p}}\log(\mathbf{\hat{p}}),
\end{matrix}
\label{equ:Conclusion of b and q}
\end{equation}
\end{footnotesize}%
where the term $\texttt{dis}[a,b]$ represents the size of the bias between a biased prediction $a$ and the true unbiased prediction $b$. The size of this bias is inversely proportional to $(-\mathbf{p}\log\mathbf{p})-(-\mathbf{\hat{p}}\log\mathbf{\hat{p}})$, and as $\texttt{dis}[\cdot,\cdot]$ increases, the information entropy~($-\mathbf{p}\log\mathbf{p}$) for category $l$ is increasingly underestimated, which requires a smaller $q^l$ for effective correction.

\input{algos}

Therefore, the class-specific parameter $q^l$ can be calculated indirectly by estimating the bias for each category. We adopt the bias estimation method from Frolic~\citep{TTA-Frolic} as follows. First, we estimate the prior probability of each class, $\tilde{\mathrm{p}}_l = \mathbf{P}(l)$, by solving the following linear equation:
\begin{equation}
    \tilde{\mathrm{p}}_{l} = \sum_{l^{'} \in \mathcal{Y}^{\text{test}}} \tilde{\mathrm{p}}_{l^{'}} \cdot \mathbb{E}_{\mathrm{x}\sim \mathbf{P}(\mathbf{x}\mid l^{'})} [ \mathbf{P}(l\mid\mathbf{x}) ].
    \label{equ:bias_estimation}
\end{equation}
A memory bank with size $M$ is maintained to store and update continuously incoming test instances for calculating $\mathbb{E}_{\mathrm{x}\sim \mathbf{P}(\mathbf{x}\mid l^{'})} [ \mathbf{P}(l\mid\mathbf{x}) ]$ in \ref{equ:bias_estimation}. Since true labels in $\mathcal{Y}^{\text{test}}$ are not available, we use pseudo-labels $\hat{l}(\mathbf{x}) \coloneqq \arg\max_l \mathbf{P}(y=l \mid \mathbf{x})$ as a substitute, following the approach Frolic~\citep{TTA-Frolic}, \textit{i.e.}:
\begin{equation}
    \mathbb{E}_{\mathrm{x}\sim \mathbf{P}(\mathbf{x}\mid l^{'})} [ \mathbf{P}(l\mid\mathbf{x}) ]
    = \frac{1}{N_{l^{'}}} \sum_{\mathbf{x} \mid \hat{l}(\mathbf{x})=l'} \mathbf{P}(l\mid\mathbf{x}),
\end{equation}
where $\mathbf{P}(l\mid\mathbf{x})$ is predicted by the model and $N_{l^{'}}=\sum_{\mathbf{x} \mid \hat{l}(\mathbf{x})=l'} 1$. We solve for $\tilde{\mathrm{p}}_{l}$ using Jacobi iteration, with a uniform initialization $\tilde{\mathrm{p}}^{(0)}_l = \frac{1}{|\mathcal{Y}^{\text{test}}|}$, $\forall l\in \mathcal{Y}^{\text{test}}$. The iteration is given by:
\begin{equation}
    \tilde{\mathrm{p}}^{(t+1)}_l =  \sum_{l^{'} \in \mathcal{Y}^{\text{test}}} \tilde{\mathrm{p}}^{(t)}_{l^{'}} \cdot \mathbb{E}_{\mathrm{x}\sim \mathbf{p}(\mathbf{x}\mid l^{'})} [ \mathbf{P}(l\mid\mathbf{x}) ].
\end{equation}
In each iteration, we perform L1 normalization over $\tilde{\mathrm{p}}^{(t)}=[\tilde{\mathrm{p}}^{(t)}_1, \dots, \tilde{\mathrm{p}}^{(t)}_{|\mathcal{Y}^{\text{test}}|}]^\top$ so that its L1 norm equals $1$. The iterative process terminates when the maximum number of iterations is reached or when the convergence condition $\|\tilde{\mathrm{p}}^{(t+1)}-\tilde{\mathrm{p}}^{(t)}\|_1 \leq \varepsilon$ is met, where $\varepsilon$ is the convergence threshold.

Combining this with Section~\ref{Correcting Biased Entropy}, we normalize the estimated bias vector $\tilde{\mathbf{p}}$ for all categories into the interval $(0,1)$. A larger bias $\tilde{\mathrm{p}}_l$ corresponds to a smaller $q^l$. We use min-max normalization to map the bias values to the parameter range:
\begin{equation}
    q^l = \alpha + (\beta-\alpha)\frac{-\log{\tilde{\mathrm{p}}_l}-\min(\tilde{\mathbf{p}})}{\max(\tilde{\mathbf{p}})-\min(\tilde{\mathbf{p}})},
    \label{equ:calcu_q}
\end{equation}
where $\alpha$ and $\beta$ define the normalization interval. After computing the class-specific parameter $q^l$, we plug it into Equation~\ref{equ:ADTE} to calculate $\mathbf{H}_\texttt{ADTE}$ for each augmented view, which is then used to select high-confidence views. After obtaining the probability distributions of the selected high-confidence views, we simply average (aggregate) these distributions to produce the final prediction. The algorithm for estimating the bias $\tilde{\mathbf{p}}$ is summarized in Algorithm~\ref{algo:estimate_b}, and the overall TTA pipeline is summarized in Algorithm~\ref{algo:pipeline}.



\section{Experiments}\label{Experiments}
\subsection{Experimental Setup}

\textbf{Benchmarks.} 
Two standard benchmarks: (a) Out-of-Distribution (OOD): evaluates generalization to distributions different from training, including ImageNet~\citep{ImageNet-1k} and variants (ImageNet-A~\citep{ImageNet-A}, -V2~\citep{ImageNet-V}, -R~\citep{ImageNet-R}, and -K~\citep{ImageNet-K}); (b) Cross-Domain: classification across diverse domains—generic objects (Caltech~\citep{Caltech}), scenes (SUN~\citep{SUN}), textures (DTD~\citep{DTD}), satellite images (EuroSAT~\citep{EuroSAT}), actions (UCF~\citep{UCF}), and five fine-grained datasets (Pets~\citep{Pets}, Cars~\citep{Cars}, Flowers~\citep{Flowers}, Food~\citep{Food}, Aircraft~\citep{Aircraft}).

\textbf{Implementation Details.}
Following~\citep{TTA-Zero,TTA-Frolic}, we use pretrained CLIP (\texttt{ViT-B/16}, \texttt{ViT-L/14}), with \texttt{ViT-B/16} serving as the default model for ablation studies. For a fair comparison, we use two types of text prompts: \textit{hand-crafted templates} from methods~\citep{TTA-TPT,TTA-Zero,TTA-TDA,TTA-BCA}, and \textit{text descriptions generated by GPTs}, as in CuPL~\citep{CuPL,TTA-Frolic}. We do not tune any hyperparameters, instead adopting the setup from Zero~\citep{TTA-Zero}, with $N=64$ augmented views and a confidence-based filtering ratio of $0.1$. The memory bank size for each category is set to $10$. The normalization interval for the class-specific parameter $q^l$ is $[0.01, 0.9]$. All experiments were conducted on a single \texttt{NVIDIA A100} GPU, with results averaged over $3$ seeds.

\subsection{Comparisons with State-of-the-art}
\begin{table}[htbp]
\captionsetup{font={footnotesize}}
\caption{Accuracy comparison (\%) on ImageNet and its variants for CLIP \texttt{ViT-B/16} and \texttt{ViT-L/14}.}
\vspace{-1em}
\label{tab:imagenet_shift}
\centering
\small
\renewcommand{\arraystretch}{1} 
\setlength{\tabcolsep}{8pt} 
\begin{tabular}{cl|ccccc|ccc}
\toprule
\multicolumn{1}{l}{} & \textbf{Method} & \textbf{IN} & \textbf{IN-V2} & \textbf{IN-K} & \textbf{IN-A} & \textbf{IN-R} & \textbf{Average} & \textbf{OOD Avg} \\ \midrule
\multicolumn{1}{l|}{} & CLIP${}_{\textcolor{orange}{\rm ~[\text{ICML 2022}]}}$ & 68.7 & 62.2 & 48.3 & 50.6 & 77.7 & 61.5 & 59.7 \\
\multicolumn{1}{l|}{} & TPT${}_{\textcolor{ orange}{\rm ~[\text{NeurIPS 2022}]}}$ & 68.9 & 63.4 & 47.9 & 54.7 & 77.0 & 62.4 & 60.8 \\ 
\multicolumn{1}{l|}{} & TDA${}_{\textcolor{ orange}{\rm ~[\text{CVPR 2024}]}}$ & 69.5 & 64.6 & 50.5 & 60.1 & 80.2 & 65.0 & 63.9 \\ 
\multicolumn{1}{l|}{} & Zero${}_{\textcolor{ orange}{\rm ~[\text{NeurIPS 2024}]}}$ & \underline{70.9} & \underline{65.1} & 50.3 & \underline{64.0} & \underline{80.8} & 
\underline{66.2} & \underline{65.0} \\
\multicolumn{1}{l|}{} & Dyna${}_{\textcolor{ orange}{\rm ~[\text{ICLR 2025}]}}$ & 69.6 & 64.7 & 48.2 & 56.2 & 78.2 & 63.4 & 61.8 \\
\multicolumn{1}{l|}{} & BCA${}_{\textcolor{ orange}{\rm ~[\text{CVPR 2025}]}}$ & 70.2 & 64.9 & \underline{50.9} & 61.1 & 80.7 & 65.6 & 64.4 \\
\multicolumn{1}{l|}{\multirow{-5}{*}{\rotatebox{90}{\footnotesize{ViT-B/16}}}} & \textbf{$\text{ADTE}_{\texttt{Templates}}$} & \textbf{71.8} & \textbf{65.6} & \textbf{53.5} & \textbf{65.5} & \textbf{81.4} & \textbf{67.5} & \textbf{66.5} \\ \cmidrule{2-9}
\multicolumn{1}{l|}{} & CuPL${}_{\textcolor{ orange}{\rm ~[\text{ICCV 2023}]}}$ & 69.9 & 64.4 & 49.4 & 59.7 & 79.5 & 64.6 & 63.3 \\
\multicolumn{1}{l|}{} & Frolic${}_{\textcolor{ orange}{\rm ~[\text{NeurIPS 2024}]}}$ & \underline{70.9} & \underline{64.7} & \underline{53.3} & \underline{60.4} & \underline{80.7} & \underline{66.0} & \underline{64.8} \\
\multicolumn{1}{l|}{} & \textbf{$\text{ADTE}_{\texttt{CuPL}}$} & \textbf{72.7} & \textbf{66.2} & \textbf{54.3} & \textbf{63.5} &\textbf{80.9} & \textbf{67.5} &\textbf{66.2} \\ \midrule[0.7pt]
\toprule[0.7pt]
\multicolumn{1}{l|}{} & CLIP${}_{\textcolor{ orange}{\rm ~[\text{ICML 2022}]}}$ & 75.9 & 70.2 & 59.7 & 70.9 & 87.9 & 72.9 & 72.2 \\
\multicolumn{1}{l|}{} & TPT${}_{\textcolor{ orange}{\rm ~[\text{NeurIPS 2022}]}}$ & 75.5 & 70.0 & 59.8 & 74.7 & 87.9 & 73.6 & 73.1 \\
\multicolumn{1}{l|}{} & TDA${}_{\textcolor{ orange}{\rm ~[\text{CVPR 2024}]}}$ & 76.3 & 71.5 & \underline{61.3} & 77.9 & 89.8 & 75.4 & 75.1 \\
\multicolumn{1}{l|}{} & Zero${}_{\textcolor{ orange}{\rm ~[\text{NeurIPS 2024}]}}$ & \underline{77.2} & \underline{71.9} & 61.1 & \underline{80.7} & \underline{90.2} & \underline{76.2} & \underline{75.9} \\
\multicolumn{1}{l|}{} & \textbf{$\text{ADTE}_{\texttt{Templates}}$} & \textbf{77.8} & \textbf{72.8} & \textbf{63.5} & \textbf{81.1} & \textbf{90.6} & \textbf{77.2} & \textbf{77.0} \\  \cmidrule{2-9}
\multicolumn{1}{l|}{\multirow{-6}{*}{\rotatebox{90}{\footnotesize{ViT-L/14}}}} & CuPL${}_{\textcolor{ orange}{\rm ~[\text{ICCV 2023}]}}$ & 76.2 & 71.9 & 60.7 & 77.9 & 89.6 & 75.3 & 75.0 \\
\multicolumn{1}{l|}{} & Frolic${}_{\textcolor{ orange}{\rm ~[\text{NeurIPS 2024}]}}$ & \underline{77.4} & \underline{72.5} & \underline{63.1} & \underline{78.9} & \underline{90.3} & \underline{76.4} & \underline{76.2} \\
\multicolumn{1}{l|}{} & \textbf{$\text{ADTE}_{\texttt{CuPL}}$} & \textbf{78.2} & \textbf{73.3} & \textbf{63.9} & \textbf{81.0} & \textbf{90.4} & \textbf{77.4} &\textbf{77.2} \\
\bottomrule
\end{tabular}
\vspace{-1em}
\end{table}
We compare \texttt{ADTE} with several state-of-the-art methods on both OOD and cross-domain benchmarks, including CLIP~\citep{VLMs-Openai-CLIP}, TPT~\citep{TTA-TPT}, TDA~\citep{TTA-TDA}, Zero~\citep{TTA-Zero}, Dyna~\citep{TTA-DynaPrompt}, BCA~\citep{TTA-BCA}, CuPL~\citep{CuPL} and Frolic~\citep{TTA-Frolic}. We present results for \texttt{ADTE} using both types of text prompts.

\textbf{Results on the OOD Benchmark}. In Table~\ref{tab:imagenet_shift}, \texttt{ADTE} consistently outperforms other methods on both \texttt{ViT-B/16} and \texttt{ViT-L/14} backbones, and with both template-based and text-description-based prompts. On ImageNet-1k, \texttt{ADTE}${}_\texttt{Templates}$ achieves $71.8\%$ accuracy, $0.9\%$ higher than Zero. \texttt{ADTE}${}_\texttt{CuPL}$ reaches $72.7\%$, outperforming Frolic by $1.8\%$. For OOD datasets, \texttt{ADTE}${}_\texttt{Templates}$ surpasses state-of-the-art methods on IN-V2~($65.6\%$ \texttt{vs} Zero's $65.1\%$), IN-K~($53.5\%$ \texttt{vs} BCA's $50.9\%$), IN-A~($65.5\%$ \texttt{vs} Zero's $64.0\%$), and IN-R~($81.4\%$ \texttt{vs} Zero's $80.8\%$). \texttt{ADTE}${}_\texttt{CuPL}$ also consistently outperforms Frolic on all datasets, with a significant $3.1\%$ lead on IN-A. For the \texttt{ViT-L/14} backbone, \texttt{ADTE}${}_\texttt{Templates}$ achieves $77.2\%$ average accuracy and $77.0\%$ OOD average accuracy, both $1.1\%$ higher than Zero. Compared to Frolic, \texttt{ADTE}${}_\texttt{CuPL}$ shows the best performance with $77.4\%$ overall average and $77.2\%$ OOD average, confirming its consistent superiority.

\begin{table}[htbp]
\captionsetup{font={footnotesize}}
\caption{Accuracy comparison (\%) on 10 cross-domain datasets for CLIP \texttt{ViT-B/16} and \texttt{ViT-L/14}.}
\label{tab:cross_data}
\vspace{-1em}
\centering
\small
\renewcommand{\arraystretch}{1.1} 
\setlength{\tabcolsep}{4.7pt} 
\begin{tabular}{cl|cccccccccc|cc}
\toprule
\multicolumn{1}{l}{} & \multirow{-2.5}{*}{\textbf{Method}} & \rotatebox{60}{\textbf{Pets.}} & \rotatebox{60}{\textbf{Flow.}} & \rotatebox{60}{\textbf{Airc.}} & \rotatebox{60}{\textbf{DTD.}} & \rotatebox{60}{\textbf{Euro.}} & \rotatebox{60}{\textbf{Cars.}} & \rotatebox{60}{\textbf{Food.}} & \rotatebox{60}{\textbf{SUN.}} & \rotatebox{60}{\textbf{Calt.}} & \rotatebox{60}{\textbf{UCF.}} & \multirow{-2.5}{*}{\textbf{Avg.}} \\ \midrule
\multicolumn{1}{l|}{} & CLIP${}_{\textcolor{orange}{\rm ~[\text{ICML 2022}]}}$ & 88.9 & 70.4 & 24.8 & 44.3 & 47.7 & 65.2 & 86.1 & 62.5 & 92.9 & 66.7 & 64.9 \\
\multicolumn{1}{l|}{} & TPT${}_{\textcolor{ orange}{\rm ~[\text{NeurIPS 2022}]}}$ & 87.7 & 68.9 & 24.7 & 47.7 & 42.4 & 66.8 & 84.6 & 65.5 & 94.1 & 68.0 & 65.0 \\ 
\multicolumn{1}{l|}{} & TDA${}_{\textcolor{ orange}{\rm ~[\text{CVPR 2024}]}}$ & 88.6 & 71.4 & 23.9 & 47.4 & \textbf{58.0} & 67.2 & \underline{86.1} & 67.6 & 94.2 & \underline{70.6} & 67.5 \\ 
\multicolumn{1}{l|}{} & Zero${}_{\textcolor{ orange}{\rm ~[\text{NeurIPS 2024}]}}$ & 87.2 & 66.8 & 24.4 & 45.9 & 43.8 & \underline{68.5} & 84.6 & 66.9 & 94.1 & 68.6 & 65.1 \\ 
\multicolumn{1}{l|}{} & Dyna${}_{\textcolor{ orange}{\rm ~[\text{ICLR 2025}]}}$ & 88.3 & 69.9 & 24.3 & 48.0 & 42.3 & 67.7 & 85.4 & 66.3 & 94.3 & 68.7 & 65.5 \\ 
\multicolumn{1}{l|}{} & BCA${}_{\textcolor{ orange}{\rm ~[\text{CVPR 2025}]}}$ & \textbf{90.4} & \textbf{73.1} & 28.6 & \textbf{53.5} & \underline{56.6} & 66.8 & 86.0 & \underline{68.4} & \underline{94.7} & 67.6 & \underline{68.6} \\ 
\multicolumn{1}{l|}{} & \textbf{$\text{ADTE}_{\texttt{Templates}}$} & \underline{89.7} & \underline{72.6} & \textbf{28.9} & \underline{49.5} & 53.8 & \textbf{70.9} & \textbf{86.3} & \textbf{70.4} & \textbf{94.8} & \textbf{73.1} & \textbf{69.0} \\ 
\cmidrule{2-13}
\multicolumn{1}{l|}{} & CuPL${}_{\textcolor{ orange}{\rm ~[\text{ICCV 2023}]}}$ & 92.0 & 73.2 & 27.7 & 54.3 & 52.7 & 66.4 & 86.2 & 68.5 & 94.6 & 70.7 & 68.6 \\ 
\multicolumn{1}{l|}{} & Frolic${}_{\textcolor{ orange}{\rm ~[\text{NeurIPS 2024}]}}$ & \textbf{92.9} & \underline{74.8} & \underline{31.5} & \underline{56.1} & \textbf{58.5} & \underline{69.1} & \textbf{87.2} & \underline{70.8} & \underline{95.2} & \underline{75.2} & \underline{71.1} \\ 
\multicolumn{1}{l|}{\multirow{-11}{*}{\rotatebox{90}{\footnotesize{ViT-B/16}}}} & \textbf{$\text{ADTE}_{\texttt{CuPL}}$} & \underline{92.7} & \textbf{75.4} & \textbf{33.5} & \textbf{57.2} & \underline{58.2} & \textbf{71.2} & \underline{86.9} & \textbf{71.3} & \textbf{95.7} & \textbf{75.5} & \textbf{71.8} \\ 
\midrule[0.7pt]
\toprule[0.7pt]
\multicolumn{1}{l|}{} & CLIP${}_{\textcolor{orange}{\rm ~[\text{ICML 2022}]}}$ & 93.5 & 79.3 & 32.4 & 53.0 & \underline{58.0} & 76.8 & 91.0 & 67.5 & 94.8 & 74.2 & 72.0 \\
\multicolumn{1}{l|}{} & TPT${}_{\textcolor{ orange}{\rm ~[\text{NeurIPS 2022}]}}$ & \underline{93.6} & 76.2 & 31.9 & 55.2 &  51.8 & 77.7 & 88.9 & 70.2 & 95.5 & 74.9 & 71.5 \\ 
\multicolumn{1}{l|}{} & TDA${}_{\textcolor{ orange}{\rm ~[\text{CVPR 2024}]}}$ & 93.5 & \textbf{80.5} & \underline{34.7} & \underline{56.7} & \textbf{64.1} & 78.3 & \underline{90.9} & 71.5 & 95.9 & 76.6 & \underline{74.2} \\ 
\multicolumn{1}{l|}{} & Zero${}_{\textcolor{ orange}{\rm ~[\text{NeurIPS 2024}]}}$ & 93.4 & 79.2 & 33.9 & 53.7 & 54.1 & \underline{78.5} & 90.2 & \underline{72.1} & \underline{96.0} & \underline{77.1} & 72.8 \\ 
\multicolumn{1}{l|}{} & \textbf{$\text{ADTE}_{\texttt{Templates}}$} & \textbf{93.7} & \underline{79.4} & \textbf{37.2} & \textbf{59.7} & 56.2 & \textbf{79.4} & \textbf{91.8} & \textbf{73.2} & \textbf{96.4} & \textbf{80.6} & \textbf{74.8} \\ \cmidrule{2-13}
\multicolumn{1}{l|}{} & CuPL${}_{\textcolor{ orange}{\rm ~[\text{ICCV 2023}]}}$ & 94.3 & 79.8 & 35.5 & 62.7 & 61.2 & 78.0 & 91.3 & 72.4 & 96.7 & 75.9 & 74.7 \\
\multicolumn{1}{l|}{} & Frolic${}_{\textcolor{ orange}{\rm ~[\text{NeurIPS 2024}]}}$ & \textbf{94.9} & \underline{82.4} & \underline{40.0} & \underline{64.1} & \textbf{66.2} & \underline{80.8} & \textbf{91.8} &\underline{74.5} & \underline{97.2} & \underline{80.0} & \underline{77.1} \\
\multicolumn{1}{l|}{\multirow{-9}{*}{\rotatebox{90}{\footnotesize{ViT-L/14}}}} &\textbf{$\text{ADTE}_{\texttt{CuPL}}$} & \underline{94.6} & \textbf{83.8} & \textbf{40.8} & \textbf{65.6} & \underline{65.9} & \textbf{81.8} & \underline{91.7} & \textbf{74.8} & \textbf{97.4} & \textbf{80.4} & \textbf{77.7} \\ 
\bottomrule
\end{tabular}
\end{table}

\textbf{Results on the Cross Domain Benchmark}. Table~\ref{tab:cross_data} demonstrates the significant advantage of \texttt{ADTE} in cross-domain image recognition. On the \texttt{ViT-B/16} backbone, \texttt{ADTE}${}_\texttt{Templates}$~($69.0\%$) outperforms all template-based methods, while \texttt{ADTE}${}_\texttt{CuPL}$~($71.8\%$) surpasses Frolic~($71.1\%$). Similarly, on \texttt{ViT-L/14}, \texttt{ADTE}${}_\texttt{Templates}$~($74.8\%$) and \texttt{ADTE}${}_\texttt{CuPL}$~($77.7\%$) both significantly outperform their respective counterparts, TDA~($74.2\%$) and Frolic~($77.1\%$). \texttt{ADTE}${}_\texttt{Templates}$ also ranks at or near the top on multiple individual datasets. For instance, on \texttt{ViT-B/16}, it improves performance by $2.4\%$, $2.0\%$, and $2.5\%$ on the Cars, SUN, and UCF datasets, respectively, compared to the best competing method. On \texttt{ViT-L/14}, it achieves state-of-the-art results on Aircraft~($37.2\%$, $+2.5\%$ improvement over TDA), DTD~($59.7\%$, $+3.0\%$ improvement over TDA), and UCF~($80.6\%$, $+3.5\%$ improvement over Zero). 


\subsection{Ablation Studies}
\begin{table}[htbp]
\centering
\small
\captionsetup{font={footnotesize}}
\caption{Accuracy~(\%) of different models on 10-datasets, including ImageNet and its five variant datasets.}
\label{tab:ablation study}
\vspace{-1em}
\renewcommand{\arraystretch}{1} 
\setlength{\tabcolsep}{5pt} 
\begin{tabular}{ll|ccc|ccc}
\toprule
& & \multicolumn{3}{c|}{ViT-B/16} & \multicolumn{3}{c}{ViT-L/14} \\  
&\multirow{-2}{*}{\textbf{Module}} & IN-Variants  & IN & 10-datasets & IN-Variants  & IN & 10-datasets \\  \midrule
$\blacktriangle$ &\textbf{$\text{ADTE}_{\texttt{Templates}}$} & \textbf{66.5} & \textbf{71.8} & \textbf{69.0} & \textbf{77.0} & \textbf{77.8} & \textbf{74.8} \\ \midrule
$\vartriangle$ &\textit{w/o} \textbf{\footnotesize ADTE} & 65.4 & 71.1 & 66.9 & 76.3 & 77.3 & 73.2 \\ 
$\vartriangle$ &\textit{w/o} \textbf{\footnotesize Logit Adjustment~(LA)} & 66.1 & 71.6 & 68.8 & 76.7 & 77.6 & 74.1 \\ \midrule
$\vartriangle$ &\textit{w/o} \textbf{\footnotesize ADTE} and \textbf{\footnotesize LA} & 65.0 & 70.9 & 65.1 & 75.9 & 77.2 & 72.8 \\ 
\bottomrule
\end{tabular}
\end{table}
\vspace{-0.5em}
We evaluate the contributions of our proposed components, \texttt{ADTE} and Logit Adjustment~(\texttt{LA}), through an ablation study. All experiments in this section are conducted on both \texttt{ViT-B/16} and \texttt{ViT-L/14} architectures, with results shown in Table~\ref{tab:ablation study}.

\textbf{Effectiveness of ADTE.}
Removing \texttt{ADTE} leads to a notable performance decline across all metrics compared to the full \texttt{ADTE}${}_\texttt{Templates}$ model. For \texttt{ViT-B/16}, accuracy drops by $1.1\%$ on IN-Variants, $0.7\%$ on IN, and $2.1\%$ on the 10-datasets. For \texttt{ViT-L/14}, accuracy decreases by $0.7\%$, $0.5\%$, and $1.6\%$ on the same benchmarks. The performance drop is most significant on the 10-datasets~($-2.1\%$ and $-1.6\%$), indicating that \texttt{ADTE} is particularly effective at enhancing the model's generalization across diverse cross-domain tasks.

\textbf{Contribution of Logit Adjustment.} \texttt{LA} adjusts the model logits by using bias estimation~\citep{TTA-Frolic}. The experiments show that removing \texttt{LA} also decreases performance, but this decline is generally smaller than removing \texttt{ADTE}. For \texttt{ViT-B/16}, the accuracy drops are $0.4\%$, $0.2\%$, and $0.2\%$, while \texttt{ViT-L/14} experiences decreases of $0.3\%$, $0.2\%$, and $0.7\%$. Both \texttt{ADTE} and \texttt{LA} contribute positively to the model's accuracy. Furthermore, \texttt{LA} can be seamlessly integrated with \texttt{ADTE} to boost performance without adding extra computational cost.

\subsection{Further Analysis}

\begin{wraptable}{r}{6cm}
\centering
\small
\vspace{-1.5em}
\captionsetup{font={footnotesize}}
\caption{Results for \texttt{SE}, \texttt{TE}, and \texttt{ADTE}.}
\vspace{-1em}
\renewcommand{\arraystretch}{0.5} 
\setlength{\tabcolsep}{1pt} 
\begin{tabular}{l|ccccc}
\toprule
\textbf{Module} & \textbf{IN} & \textbf{IN-V2} & \textbf{IN-K} & \textbf{IN-A} & \textbf{IN-R} \\ \midrule
LA-SE & 70.7 & 64.1 & 52.3 & 64.0 & 80.2 \\
SE-LA & 71.0 & 64.3 & 52.5 & 64.1 & 80.3 \\
TE-LA~(q=0.5) & 71.6 & 65.2 & 53.0 & 64.8 & 80.7 \\ \midrule
\textbf{ADTE} & \textbf{71.8} & \textbf{65.6} & \textbf{53.5} & \textbf{65.5} & \textbf{81.4} \\
\bottomrule
\end{tabular}
\label{tab:Comparison among SE, TE, and ADTE}
\vspace{-1em}
\end{wraptable}
\textbf{Comparison Among \texttt{SE}, \texttt{TE}, and \texttt{ADTE}}.
Table~\ref{tab:Comparison among SE, TE, and ADTE} compares the performance of \texttt{ADTE}, \texttt{SE}, and \texttt{TE}, all with logit adjustment~(LA). \texttt{ADTE} demonstrates the highest performance. This highlights \texttt{ADTE}'s distinct advantages in tasks that rely on entropy information.

\begin{figure}[htbp]
\centering
\vspace{-0.5em}
\begin{minipage}{.55\textwidth}
    \centering
    \small
    \captionsetup{font={footnotesize}}
    \captionof{table}{Computational cost and effect of different intervals.}
    \vspace{-1em}
    \renewcommand{\arraystretch}{1} 
    \setlength{\tabcolsep}{4pt} 
    \begin{tabular}{l|cccccc}
    \toprule
    \textbf{Interval} & \textbf{IN} & \textbf{IN-V2} & \textbf{IN-K} & \textbf{IN-A} & \textbf{IN-R} &\textbf{Avg} \\ \midrule
    ${[0.1, 0.9]}$ & 71.7 & 65.5 & 53.4 & 65.2 & 81.3 & 67.4 \\
    $\mathbf{[0.01, 0.9]}$ & \textbf{71.8} & \textbf{65.6} & 53.5 & 65.5 & \textbf{81.4} & \textbf{67.5} \\
    $[0.001, 0.9]$ & \textbf{71.8} & \textbf{65.6} & \textbf{53.6} & \textbf{65.6} & 81.2 & \textbf{67.5} \\
    $[0.0001, 0.9]$ & 71.7 & 65.5 & \textbf{53.6} & \textbf{65.6} & 81.2 & \textbf{67.5} \\
    \bottomrule
    \toprule
    \textbf{Metric} & \multicolumn{2}{c}{TPT} & \multicolumn{2}{c}{Zero} & \multicolumn{2}{c}{\textbf{ADTE}} \\ \midrule
    Time[$s$] & \multicolumn{2}{c}{0.57±0.01} & \multicolumn{2}{c}{0.06±0.01} & \multicolumn{2}{c}{0.07±0.01} \\
    Mem[GB] & \multicolumn{2}{c}{17.66} & \multicolumn{2}{c}{1.40} & \multicolumn{2}{c}{1.46} \\
    \bottomrule
    \end{tabular}
    \vspace{-0.5em}
 \label{tab:interval}
\end{minipage}%
\hfill
\begin{minipage}{.43\textwidth}
    \centering
    \includegraphics[width=1\textwidth]{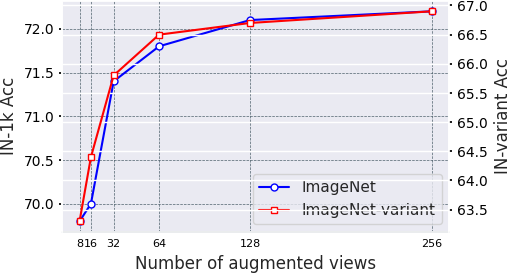}
    \vspace{-1.5em}
    \captionsetup{font={footnotesize}}
    \caption{Different number of views.}
    \vspace{-4em}
    \label{fig:augmented_views}
\end{minipage}
\end{figure}

\textbf{Normalization Intervals and Computational Cost}. Table~\ref{tab:interval} shows the performance \texttt{ADTE} using different normalization intervals. As the lower bound of the interval decreases from $0.1$ to $0.0001$, performance generally improves and then stabilizes. This demonstrates that \texttt{ADTE} is effective when the class-specific parameter is less than $1$ and that the method is robust as it does not require precise parameter tuning; as long as the lower bound is sufficiently small, excellent performance is achieved. Furthermore, \texttt{ADTE} adds only a negligible computational cost compared to Zero, primarily for the bias estimation step.

\textbf{Number of Augmented Views}. Figure~\ref{fig:augmented_views} shows that increasing views ($N$) improves performance until convergence. And more views help \texttt{ADTE} select more confident views, boosting accuracy.

\textbf{Memory Bank Size.}
We systematically evaluated the impact of varying memory bank sizes on \texttt{ADTE}'s performance on ImageNet-1k, with quantitative results presented in Table~\ref{tab:memory_bank_size}.

\begin{table}[htbp]
\vspace{-0.5em}
\centering
\small
\captionsetup{font={footnotesize}}
\caption{Impact of memory bank size on performance (\%).}
\vspace{-1em}
\renewcommand{\arraystretch}{0.8}
\setlength{\tabcolsep}{5pt}
\begin{tabular}{l|ccccccc}
\toprule
\textbf{Size} & \textbf{1} & \textbf{2} & \textbf{5} & \textbf{10} & \textbf{20} & \textbf{30} & \textbf{40} \\
\midrule
\textbf{ADTE} & 71.71 & 71.73 & 71.74 & 71.83 & 71.81 & 71.85 & 71.86 \\
\bottomrule
\end{tabular}
\label{tab:memory_bank_size}
\vspace{-0.5em}
\end{table}
The results show that the memory bank size has a minimal impact on \texttt{ADTE}'s performance: with a size of $1$, the model already achieves $71.71\%$ accuracy. As the size increases to $10$, the accuracy stabilizes around $71.83\%$. Further expansion to $40$ yields only marginal fluctuations ($71.86\%$). A small number of high-confidence samples is sufficient to support reliable adaptive adjustments.

Note that $71.71\%$ accuracy achieved with size $1$ does not mean the memory bank can be removed. To see this, consider $200$ categories, using a memory bank size of $1$ yields 200 samples in total. The statistical information used to calculate bias for a specific category is derived not only from its own sample but from all $200$ pseudo-labeled samples in the bank.

\textbf{Improvements on \texttt{tail} Categories.} To further validate \text{ADTE}'s effectiveness on 
\texttt{tail} classes, we conducted quantitative experiments on \texttt{tail}
categories in ImageNet-1k. We selected the last $10$ tail classes in ImageNet-1k and compared the performance between CLIP and \texttt{ADTE}.

\begin{table}[h!]
\vspace{-0.5em}
\centering
\small
\captionsetup{font={footnotesize}}
\caption{Performance improvement on representative tail classes.}
\vspace{-1em}
\renewcommand{\arraystretch}{1.1}
\setlength{\tabcolsep}{5pt}
\begin{tabular}{l|cccccccccc}
\toprule
\textbf{Class Index} & \textbf{670} & \textbf{193} & \textbf{981} & \textbf{157} & \textbf{533} & \textbf{168} & \textbf{316} & \textbf{106} & \textbf{50} & \textbf{156} \\
\midrule
Accuracy (CLIP) & 17.4 & 16.7 & 16.1 & 15.0 & 14.5 & 12.5 & 11.5 & 0.00 & 0.00 & 0.00 \\
Confidence (CLIP) & 0.01 & 0.11 & 0.05 & 0.18 & 0.09 & 0.13 & 0.15 & 0.09 & 0.13 & 0.19 \\
\midrule
\textbf{Accuracy (ADTE)} & \textbf{16.7} & \textbf{33.3} & \textbf{33.3} & \textbf{38.5} & \textbf{25.0} & \textbf{59.4} & \textbf{35.0} & \textbf{32.2} & \textbf{41.7} & \textbf{56.5} \\
\textbf{Confidence (ADTE)} & \textbf{0.16} & \textbf{0.31} & \textbf{0.28} & \textbf{0.35} & \textbf{0.24} & \textbf{0.44} & \textbf{0.33} & \textbf{0.25} & \textbf{0.34} & \textbf{0.47} \\
\bottomrule
\end{tabular}
\label{tab:tail_class_performance}
\end{table}
As shown in Table~\ref{tab:tail_class_performance}, \texttt{ADTE} achieves significant improvements in both accuracy and confidence for these categories. For instance, classes that originally had zero accuracy show enhanced accuracy ranging from $32.2\%$ to $56.5\%$ under \texttt{ADTE}, while confidence scores simultaneously increase. This directly validates \texttt{ADTE}'s effectiveness in improving \texttt{tail} class performance.
\begin{table}[h!]
\centering
\small
\captionsetup{font={footnotesize}}
\caption{Comparative analysis between head and tail classes.}
\vspace{-1em}
\renewcommand{\arraystretch}{1.1}
\setlength{\tabcolsep}{5pt}
\begin{tabular}{l|cccc}
\toprule
\textbf{Module} & \textbf{Head (avg\_pred)} & \textbf{Head (avg\_entropy)} & \textbf{Tail (avg\_pred)} & \textbf{Tail (avg\_entropy)} \\
\midrule
CLIP & 0.6687 & 5.2956 & 0.1466 & 5.2972 \\
\textbf{ADTE} & \textbf{0.8112} & \textbf{3.2156} & \textbf{0.3638} & \textbf{3.3761} \\
\bottomrule
\end{tabular}
\label{tab:head_tail_comparison}
\end{table}

Table~\ref{tab:head_tail_comparison} presents results of prediction confidence and entropy values for head classes (top $50$) and tail classes (bottom $50$). The results demonstrate that \texttt{ADTE} not only enhances the average prediction confidence for \texttt{tail} classes (from $0.1466$ to $0.3638$) but also reduces prediction uncertainty by decreasing the entropy value (from $5.2972$ to $3.3761$). These improvements indicate that the model achieves more definitive and reliable predictions for \texttt{tail} categories.

\begin{wraptable}{r}{6cm}
\centering
\small
\vspace{-1.5em}
\captionsetup{font={footnotesize}}
\caption{Results with different normalizations.}
\vspace{-1em}
\renewcommand{\arraystretch}{0.8} 
\setlength{\tabcolsep}{1pt} 
\begin{tabular}{l|ccccc}
\toprule
\textbf{Normalization} & \textbf{IN} & \textbf{IN-V2} & \textbf{IN-K} & \textbf{IN-A} & \textbf{IN-R} \\ \midrule
Z-Score & 71.8 & 65.9 & \textbf{53.5} & 65.5 & 81.3 \\
Decimal Scaling & \textbf{72.0} & 65.8 & \textbf{53.5} & \textbf{65.7} & 81.3 \\
Sigmoid & \textbf{72.0} & \textbf{66.1} & \textbf{53.5} & 65.4 & 81.2 \\
\textbf{Min-Max} & 71.8 & 65.6 & \textbf{53.5} & 65.5 & \textbf{81.4} \\
\bottomrule
\end{tabular}
\label{tab:Normalization functions}
\vspace{-1.5em}
\end{wraptable}
\textbf{Effect of Different Normalization Functions}. As shown in Table~\ref{tab:Normalization functions}, \texttt{ADTE}'s performance is not highly dependent on a specific normalization function. While some methods may offer minor advantages on certain datasets, no single method is universally optimal. This demonstrates the robustness of our approach.

\section{Conclusion and Limitation}\label{sec:conclusion}
\textbf{Conclusion.}
This paper introduces a novel Test-Time Adaptation~(TTA) approach called \textbf{A}daptive \textbf{D}ebiasing \textbf{T}sallis \textbf{E}ntropy~(\texttt{ADTE}), which is designed to handle the inherent prediction biases in Vision-Language Models~(VLMs). We show that Shannon Entropy can be considered a special case of Tsallis Entropy, and that its performance serves as a lower bound. By generalizing Tsallis Entropy with class-specific parameters $q^l$ tailored for each category $l$, \texttt{ADTE} effectively reduces the biases encountered during TTA. Our experimental results demonstrate that \texttt{ADTE} outperforms state-of-the-art methods across multiple datasets, proving its robustness and effectiveness in improving the adaptation performance of VLMs.


\textbf{Limitation.}
Due to the specific nature of its design, \texttt{ADTE} is highly effective in scenarios with significant prediction bias. However, its advantages may be less apparent in scenarios where the model has almost no biased predictions.

\section*{Acknowledgements}
This work is supported by the NSFC (62276131,  62506168), Natural Science Foundation of Jiangsu Province of China under Grant (BK20240081, BK20251431).

The authors gratefully acknowledge financial support from the China Scholarship Council~(CSC) (Grant No. 202506840036).

\bibliography{iclr2026_conference}
\bibliographystyle{iclr2026_conference}

\newpage
\appendix

\begin{center}
\bf\large Appendix for Adaptive Debiasing Tsallis Entropy for Test-Time Adaptation
\end{center}

\section{Additional Experiments}
\setcounter{definition}{0} 
\setcounter{property}{0} 

\textbf{Detailed Analysis of Estimated Bias Statistics.}
We conduct a statistical analysis of prediction bias on ImageNet variants and cross-domain datasets to explore the correlation between performance gains and the degree of bias. Table~\ref{tab:bias_analysis} shows results on five ImageNet variants.
\begin{table}[htbp]
\vspace{-0.5em}
\centering
\small
\captionsetup{font={footnotesize}}
\caption{Bias statistical analysis of ImageNet variant datasets.}
\vspace{-1em}
\renewcommand{\arraystretch}{1}
\setlength{\tabcolsep}{5pt}
\begin{tabular}{l|ccccc}
\toprule
\textbf{Dataset (Classes)} & \textbf{IN-A (200)} & \textbf{IN-R (200)} & \textbf{IN-K (1000)} & \textbf{IN-V (1000)} & \textbf{IN-I (1000)} \\
\midrule
Variance & 0.9527 & 0.7650 & 1.5858 & 1.1603 & 1.0833 \\
Std\_Dev & 0.9621 & 0.8747 & 1.2953 & 1.0772 & 1.0408 \\
\midrule
Acc (CLIP) & 50.6 & 77.7 & 48.3 & 62.2 & 68.7 \\
\textbf{Acc (ADTE)} & \textbf{65.5} & \textbf{81.4} & \textbf{53.5} & \textbf{65.6} & \textbf{71.8} \\
Gain & 14.9 & 4.7 & 5.2 & 3.4 & 3.1 \\
\bottomrule
\end{tabular}
\label{tab:bias_analysis}
\vspace{-1.0em}
\end{table}

As shown in Table~\ref{tab:bias_analysis}, datasets with larger bias variance and standard deviation (e.g., IN-A and IN-K) correspond to a greater disparity in bias between \texttt{head} and \texttt{tail} classes. Consequently, the performance gains of \texttt{ADTE} are higher on these datasets ($14.9\%$ and $5.2\%$). This indicates that \texttt{ADTE} is particularly effective in scenarios with prominent biases, which also explains the variation in performance improvements across different datasets.

For the cross-domain benchmark, we analyzed the average prediction accuracy and confidence across different percentile intervals. As shown in Tables~\ref{tab:cross_domain_bias_part1} and \ref{tab:cross_domain_bias_part2}, all cross-domain datasets exhibit significant long-tail distribution characteristics and CLIP demonstrates pronounced prediction bias across them. These results validate the universality of CLIP's prediction bias.
\begin{table}[htbp]
\captionsetup{font={footnotesize}}
\caption{Quantitative analysis of prediction bias on cross-domain datasets (Part 1). The values in each cell represent (Accuracy, Confidence).}
\vspace{-1em}
\label{tab:cross_domain_bias_part1}
\centering
\small
\renewcommand{\arraystretch}{1.1}
\setlength{\tabcolsep}{4.7pt}
\begin{tabular}{l|ccccc}
\toprule
\textbf{Class Percentile} & \textbf{Flower102} & \textbf{DTD} & \textbf{Pets} & \textbf{Cars} & \textbf{UCF101} \\
\midrule
0\%-10\% & (1.0, 0.85) & (0.98, 0.82) & (1.0, 0.95) & (0.99, 0.88) & (1.0, 0.95) \\
10\%-20\% & (1.0, 0.91) & (0.86, 0.67) & (0.99, 0.96) & (0.94, 0.81) & (0.98, 0.86) \\
20\%-30\% & (0.98, 0.86) & (0.74, 0.59) & (0.98, 0.93) & (0.89, 0.71) & (0.94, 0.80) \\
30\%-40\% & (0.93, 0.82) & (0.54, 0.32) & (0.96, 0.87) & (0.82, 0.65) & (0.90, 0.74) \\
40\%-50\% & (0.87, 0.69) & (0.42, 0.29) & (0.95, 0.87) & (0.73, 0.51) & (0.82, 0.63) \\
50\%-60\% & (0.81, 0.63) & (0.34, 0.23) & (0.93, 0.80) & (0.64, 0.44) & (0.72, 0.55) \\
60\%-70\% & (0.62, 0.45) & (0.22, 0.16) & (0.90, 0.81) & (0.56, 0.38) & (0.57, 0.41) \\
70\%-80\% & (0.50, 0.35) & (0.07, 0.08) & (0.81, 0.67) & (0.42, 0.34) & (0.32, 0.27) \\
80\%-90\% & (0.12, 0.14) & (0.01, 0.04) & (0.63, 0.52) & (0.21, 0.22) & (0.19, 0.19) \\
90\%-100\% & (0.00, 0.01) & (0.00, 0.03) & (0.00, 0.00) & (0.06, 0.11) & (0.06, 0.08) \\
\bottomrule
\end{tabular}
\end{table}

\begin{table}[htbp]
\captionsetup{font={footnotesize}}
\caption{Quantitative analysis of prediction bias on cross-domain datasets (Part 2). The values in each cell represent (Accuracy, Confidence).}
\vspace{-1em}
\label{tab:cross_domain_bias_part2}
\centering
\small
\renewcommand{\arraystretch}{1.1}
\setlength{\tabcolsep}{4.7pt}
\begin{tabular}{l|ccccc}
\toprule
\textbf{Class Percentile} & \textbf{Caltech101} & \textbf{Food101} & \textbf{SUN397} & \textbf{Aircraft} & \textbf{Eurosat} \\
\midrule
0\%-10\% & (1.0, 0.96) & (0.96, 0.93) & (0.96, 0.84) & (0.86, 0.66) & (0.82, 0.51) \\
10\%-20\% & (1.0, 0.96) & (0.94, 0.87) & (0.90, 0.75) & (0.53, 0.24) & (0.77, 0.63) \\
20\%-30\% & (1.0, 0.97) & (0.93, 0.84) & (0.85, 0.69) & (0.37, 0.14) & (0.76, 0.57) \\
30\%-40\% & (1.0, 0.97) & (0.91, 0.85) & (0.79, 0.60) & (0.24, 0.12) & (0.72, 0.43) \\
40\%-50\% & (1.0, 0.96) & (0.89, 0.82) & (0.72, 0.52) & (0.17, 0.12) & (0.47, 0.35) \\
50\%-60\% & (0.98, 0.90) & (0.87, 0.79) & (0.64, 0.47) & (0.10, 0.12) & (0.29, 0.32) \\
60\%-70\% & (0.93, 0.87) & (0.85, 0.76) & (0.56, 0.39) & (0.07, 0.07) & (0.15, 0.24) \\
70\%-80\% & (0.88, 0.79) & (0.82, 0.72) & (0.47, 0.32) & (0.04, 0.07) & (0.14, 0.13) \\
80\%-90\% & (0.81, 0.72) & (0.76, 0.65) & (0.32, 0.24) & (0.00, 0.04) & (0.00, 0.08) \\
90\%-100\% & (0.49, 0.42) & (0.64, 0.54) & (0.12, 0.12) & (0.00, 0.03) & (0.00, 0.04) \\
\bottomrule
\end{tabular}
\end{table}

\textbf{Correlation between $\mathbf{Tcr}_K$ and Prediction Accuracy.} We examine the correlation between $\mathbf{Tcr}_K$ and prediction accuracy on ImageNet-A to validate the reliability of $\mathbf{Tcr}_K$ as an intermediate evaluation metric.
\begin{table}[htbp]
\centering
\small
\captionsetup{font={footnotesize}}
\caption{Correlation between $\mathbf{Tcr}_K$ and accuracy on ImageNet-A.}
\vspace{-1em}
\renewcommand{\arraystretch}{1.1}
\setlength{\tabcolsep}{5pt}
\begin{tabular}{l|cccccc}
\toprule
\textbf{Value of $q$} & \textbf{$\mathbf{Tcr}_1$} & \textbf{$\mathbf{Tcr}_3$} & \textbf{$\mathbf{Tcr}_5$} & \textbf{$\mathbf{Tcr}_{10}$} & \textbf{$\mathbf{Tcr}_{20}$} & \textbf{Accuracy} \\
\midrule
0.01 & 29.3069 & 27.0006 & 25.9538 & 24.6061 & 23.7410 & 63.48 \\
0.1 & 29.3283 & 26.9548 & 25.9067 & 24.5704 & 23.7164 & 63.45 \\
0.5 & 29.3293 & 26.8951 & 25.8540 & 24.5383 & 23.6985 & 63.24 \\
0.9 & 29.3189 & 26.8414 & 25.8114 & 24.5162 & 23.6892 & 62.80 \\
1.1 & 29.3043 & 26.8000 & 25.7808 & 24.5020 & 23.6841 & 62.58 \\
1.5 & 29.2820 & 26.7511 & 25.7460 & 24.4864 & 23.6785 & 62.68 \\
2.0 & 29.2682 & 26.7257 & 25.7291 & 24.4798 & 23.6771 & 60.95 \\
\bottomrule
\end{tabular}
\label{tab:tcrk_accuracy_correlation}
\end{table}

As shown in Table~\ref{tab:tcrk_accuracy_correlation}, when keeping $K$ constant while decreasing $q$, both $\mathbf{Tcr}_K$ and accuracy demonstrate an increasing trend. This suggests that $\mathbf{Tcr}_K$ and accuracy can be considered approximately equivalent metrics, as the former evaluates model performance through prediction confidence, while the latter directly measures performance via top-1 prediction correctness.

\textbf{Integration with DEYO Method.} We further investigate the modularity of \texttt{ADTE} by integrating it with DEYO \citep{lee2024entropy}. By simply replacing the Softmax Entropy used in DEYO with our \texttt{ADTE} while keeping all other parameters and pretrained weights unchanged, we achieve significant performance improvements.
\begin{table}[htbp]
\centering
\small
\captionsetup{font={footnotesize}}
\caption{Integration with DEYO method on ColoredMNIST.}
\vspace{-1em}
\renewcommand{\arraystretch}{1.1}
\setlength{\tabcolsep}{5pt}
\begin{tabular}{l|ccc}
\toprule
\textbf{Method} & \textbf{Avg Acc (\%)} & \textbf{Worst-Group Acc (\%)} & \textbf{Time (s)} \\
\midrule
DEYO & 78.24 & 67.39 & 0.073 \\
DEYO + ADTE & 79.29 & 69.65 & 0.074 \\
\bottomrule
\end{tabular}
\label{tab:deyo_integration}
\end{table}
Table~\ref{tab:deyo_integration} demonstrates that the modified method obtains a $2.26\%$ boost in Worst-Group Accuracy with almost no additional computational overhead.

\textbf{Estimated Bias Stability.} We tracked the changes in the estimated bias's mean and variance as the size of the memory bank increases, with results presented in Table~\ref{tab:bias_stability}.
\begin{table}[h!]
\centering
\small
\captionsetup{font={footnotesize}}
\caption{Stability of estimated bias over memory bank size on ImageNet-1k.}
\vspace{-1em}
\renewcommand{\arraystretch}{1.1}
\setlength{\tabcolsep}{5pt}
\begin{tabular}{l|ccccccc}
\toprule
\textbf{Size} & \textbf{1} & \textbf{2} & \textbf{5} & \textbf{10} & \textbf{20} & \textbf{30} & \textbf{40} \\
\midrule
Mean & -5.2248 & -4.3044 & -4.2567 & -3.8658 & -3.9067 & -3.8224 & -3.8825 \\
Variance & 2.2897 & 2.0415 & 1.7100 & 1.5247 & 1.4618 & 1.4537 & 1.4752 \\
\bottomrule
\end{tabular}
\label{tab:bias_stability}
\end{table}
As the size of the memory bank increases, the variance of the estimated biases gradually decreases. When the size reaches $10$, both the mean and the variance tend to stabilize, indicating that only $10$ samples per category are needed to reach a stable bias estimation. This trend is consistent with the results of the performance experiment, where the performance also stabilizes around the size of $10$. \texttt{ADTE} can quickly converge to a reliable bias estimation during the test without relying on an excessively large memory bank.

\textbf{Experiments on Corrupted Datasets (ImageNet-C, CIFAR-10-C, CIFAR-100-C).} To further demonstrate the effectiveness of CGPO, we conducted experiments on ImageNet-C, CIFAR-10-C, and CIFAR-100-C datasets~\citep{hendrycks2019benchmarking}. For ImageNet-C, we randomly select $3$ representative corruption families (Defocus Blur,
Glass Blur, Motion Blur). For CIFAR-10-C and CIFAR-100-C, we evaluated the models on $7$ diverse corruption types, covering blur, noise, and geometric distortions.

As shown in Tables~\ref{tab:imagenet_c}, \ref{tab:cifar10_c} and \ref{tab:cifar100_c}, across all datasets and nearly all corruption types, ADTE consistently outperforms both the original CLIP and SE. Importantly, (1) SE often degrades performance
compared to the CLIP model, especially under severe blur and noise; (2) ADTE remains stable
and delivers consistent improvements, demonstrating stronger robustness even in settings where SE
becomes unreliable.

\begin{table}[h!]
\centering
\caption{{Accuracy (\%) on ImageNet-C across severity levels (1--5).}}
\vspace{-1em}
\label{tab:imagenet_c}
\resizebox{\textwidth}{!}{%
    \setlength{\tabcolsep}{2pt} 
    \begin{tabular}{@{}l rrrrr rrrrr rrrrr@{}}
    \toprule
    \textbf{Model} & \multicolumn{5}{c}{\textbf{defocus\_blur}} & \multicolumn{5}{c}{\textbf{glass\_blur}} & \multicolumn{5}{c}{\textbf{motion\_blur}} \\
    \cmidrule(lr){2-6} \cmidrule(lr){7-11} \cmidrule(lr){12-16}
    & 1 & 2 & 3 & 4 & 5 & 1 & 2 & 3 & 4 & 5 & 1 & 2 & 3 & 4 & 5 \\
    \midrule
    CLIP-ViT/B-16 & 58.78 & 53.82 & 43.36 & 34.42 & 26.12 & 56.47 & 48.27 & 26.75 & 20.88 & 16.96 & 62.51 & 56.91 & 47.46 & 34.37 & 26.55 \\
    SE & 58.62 & 54.06 & 43.15 & 34.35 & 26.39 & 55.74 & 46.98 & 26.34 & 19.87 & 15.56 & 62.69 & 57.18 & 46.41 & 33.87 & 25.31 \\
    \textbf{ADTE} & \textbf{59.43} & \textbf{54.89} & \textbf{44.05} & \textbf{35.06} & \textbf{26.95} & \textbf{57.56} & \textbf{48.52} & \textbf{27.58} & \textbf{21.65} & \textbf{17.84} & \textbf{63.83} & \textbf{58.23} & \textbf{48.22} & \textbf{34.54} & \textbf{27.06} \\
    \bottomrule
    \end{tabular}%
}
\vspace{-1em}
\end{table}

\begin{table}[h!]
\centering
\footnotesize
\caption{Average Accuracy (\%) on CIFAR-10-C.}
\vspace{-1em}
\label{tab:cifar10_c}
\resizebox{\textwidth}{!}{%
\setlength{\tabcolsep}{2pt}
\begin{tabular}{@{}l rrrrrrrr@{}}
\toprule
\textbf{Model} & \textbf{brightness} & \textbf{elastic\_transform} & \textbf{gaussian\_blur} & \textbf{impulse\_noise} & \textbf{motion\_blur} & \textbf{shot\_noise} & \textbf{speckle\_noise} \\
\midrule
CLIP-ViT/B-16 & 90.94 & 84.33 & \textbf{90.54} & 87.94 & 87.33 & 84.73 & 85.01 \\
SE & 89.86 & 83.71 & 88.92 & 86.88 & 86.98 & 84.39 & 84.86 \\
\textbf{ADTE} & \textbf{91.32} & \textbf{84.86} & \underline{89.96} & \textbf{88.43} & \textbf{87.65} & \textbf{85.56} & \textbf{85.42} \\
\bottomrule
\end{tabular}
}
\vspace{-1em}
\end{table}

\begin{table}[h!]
\centering
\footnotesize
\caption{Average Accuracy (\%) on CIFAR-100-C.}
\vspace{-1em}
\label{tab:cifar100_c}
\resizebox{\textwidth}{!}{%
\setlength{\tabcolsep}{2pt}
\begin{tabular}{@{}l rrrrrrrr@{}}
\toprule
\textbf{Model} & \textbf{brightness} & \textbf{elastic\_transform} & \textbf{gaussian\_blur} & \textbf{impulse\_noise} & \textbf{motion\_blur} & \textbf{shot\_noise} & \textbf{speckle\_noise} \\
\midrule
CLIP-ViT/B-16 & 68.29 & 57.83 & \textbf{68.05} & 62.46 & \textbf{62.17} & 57.27 & \textbf{57.66} \\
SE & 67.92 & 58.05 & 66.56 & 61.78 & 61.32 & 57.01 & 56.22 \\
\textbf{ADTE} & \textbf{68.80} & \textbf{59.32} & \underline{67.68} & \textbf{62.65} & \underline{61.98} & \textbf{57.56} & \underline{57.43} \\
\bottomrule
\end{tabular}
}
\end{table}

Consistent gains in $3$ datasets and most corruption severities indicate the generalization ability of
ADTE, reinforcing our main claim: ADTE provides a more robust correction mechanism than SE
under distribution shifts, even when SE partially or completely fails

\textbf{Analysis of Pseudo-Label Noise Effects on Class-Wise Bias Estimation.} To directly measure the effect of noisy pseudo-labels, we conducted controlled experiments in which we manually inject pseudo-label noise on the ImageNet-
A dataset. Specifically, for each sample, with probability $p \in \{20\%, 40\%, 60\%, 80\%, 100\%\}$, the true label is replaced by a random incorrect label. This allows us to isolate the impact of degraded pseudo-label quality under varying noise intensities.

\begin{table}[h!]
\centering
\footnotesize
\caption{ImageNet-A Accuracy (\%).}
\vspace{-1em}
\label{tab:imagenet_a_accuracy}
\resizebox{\textwidth}{!}{%
\setlength{\tabcolsep}{2pt}
\begin{tabular}{@{}l|c|c|c|c|c|c|c|c @{}}
\toprule
Model & \textbf{Zero-shot} & \textbf{True Label} & \textbf{Pseudo Label} & \textbf{20\% Noise} & \textbf{40\% Noise} & \textbf{60\% Noise} & \textbf{80\% Noise} & \textbf{100\% Noise} \\
\midrule
CLIP-ViT-B/16 & 50.6 & -- & -- & -- & -- & -- & -- & -- \\
\midrule
SE & 64.0 & -- & -- & -- & -- & -- & -- & --  \\
\midrule
ADTE & -- & \textbf{65.9} & 65.5 & \textbf{65.9} & 65.8 & 65.6 & 65.4 & 64.6 \\
\bottomrule
\end{tabular}
}
\end{table}

The results are shown in Table~\ref{tab:imagenet_a_accuracy}. We observe $3$ important trends: (1) When using the true label to estimate bias, ADTE presents the best performance compared with results under pseudo-label noise; (2) ADTE remains extremely stable under moderate and even severe noise levels: accuracy stays within a very narrow band ($65.9 \to 65.4$) even as pseudo-label corruption increases to 80\%; (3) Even with 100\% incorrect pseudo-labels, ADTE still outperforms SE ($64.6$ vs. $64.0$).

This aligns precisely with our theoretical analysis in Sections~\ref{Tsallis Entropy} and \ref{Correcting Biased Entropy}:
SE is a strict lower bound of TE, and TE is a strict lower bound of ADTE, since SE corresponds to the special case where all category-wise parameters $q^l$ are identical. Therefore, ADTE can never perform worse than SE, even under extreme pseudo-label noise.

Overall, ADTE is robust to pseudo-label errors, noise-insensitive, and maintains reliable performance even under worst-case degradation, making it practical for real-world deployment where pseudo-labels are inevitably imperfect.

\textbf{Performance under Low-Bias Scenarios.} To quantify the performance gap between ADTE and SE/TE under low-bias conditions, we conducted experiments on ImageNet-1k under progressively lower bias conditions. Specifically, we sort ImageNet-1k classes by their prediction-bias magnitude and randomly construct 5 subsets:
\begin{itemize}
    \item $s_1$: highest inter-class bias (200 most biased classes)
    \item $s_2$: ...
    \item $s_5$: lowest bias (200 least biased classes)
\end{itemize}
This setup directly evaluates ADTE, TE, and SE in increasingly unbiased distributions. The results in Table~\ref{tab:imagenet_a_stages} show that: (1) ADTE consistently outperforms SE and TE at all bias levels; (2) The advantage grows as the bias becomes larger; (3) Even in nearly unbiased settings, ADTE still yields notable gains. This quantitatively validates ADTE’s applicability limit and supports the theoretical claim of lower-bound.

\begin{table}[h!]
\centering
\footnotesize
\caption{ImageNet-A accuracy (\%) across progressively lower bias conditions.}
\vspace{-1em}
\label{tab:imagenet_a_stages}
\begin{tabular}{@{}l rrrrr@{}}
\toprule
\textbf{Model} & \textbf{$s_1$} & \textbf{$s_2$} & \textbf{$s_3$} & \textbf{$s_4$} & \textbf{$s_5$} \\
\midrule
CLIP-ViT/B-16 & 69.5 & 58.8 & 52.3 & 74.2 & 76.3 \\
SE & 70.4 & 60.3 & 53.1 & 75.9 & 77.8 \\
ADTE & \textbf{73.8} & \textbf{62.3} & \textbf{55.2} & \textbf{77.1} & \textbf{78.5} \\
\bottomrule
\end{tabular}
\end{table}

\textbf{Continual or Gradual Domain-Shift Scenarios.} To evaluate ADTE under realistic evolving distributions, we conducted new experiments by mixing five ImageNet variants, i.e., ImageNet-1k, ImageNet-A, ImageNet-V, ImageNet-K, and ImageNet-R, in a fully randomized interleaved stream. This simulates an online TTA scenario where the domain changes unpredictably from sample to sample, making it one of the most challenging continual shift settings.

\begin{table}[h!]
\centering
\footnotesize
\caption{Accuracy (\%) on randomized, mixed five ImageNet variants.}
\vspace{-1em}
\label{tab:mixed_five}
\begin{tabular}{@{}l rrrrr@{}}
\toprule
\textbf{Model} & \textbf{ImageNet-1k} & \textbf{ImageNet-A} & \textbf{ImageNet-V} & \textbf{ImageNet-K} & \textbf{ImageNet-R} \\
\midrule
ADTE & 71.8 & 65.5 & 65.6 & 53.5 & 81.4 \\
ADTE\_random & 72.0 & 65.8 & 65.4 & 53.5 & 81.2 \\
\bottomrule
\end{tabular}
\end{table}

The results in Table~\ref{tab:mixed_five} demonstrate: (1) ADTE consistently outperforms SE and the CLIP baseline in all domains; (2) No signs of instability or degradation, even when the domain identity changes every few samples; (3) Memory-based bias estimation remains effective because only a small class-wise bank is maintained, which is naturally robust to cross-domain mixing.

\textbf{Evaluating ADTE’s Applicability  Unimodal ImageNet Models and CLIP Successors like SigLIP.} To verify the effectiveness of ADTE on models other than CLIP, we first conducted additional experiments on the ImageNet-A dataset on unimodal ImageNet-pretrained models, following the MEMO~\citep{TTA-MEMO} test-time adaptation benchmark. As shown in Table~~\ref{tab:uni_modal}, ADTE consistently improves performance on ResNext-101, outperforming standard TTA and other adaptation techniques. The results demonstrate that ADTE is not tied to multimodal encoders, and it remains effective on purely vision-based architectures.

\begin{table}[h!]
\centering
\footnotesize
\caption{ImageNet-A Error (\%).}
\vspace{-1em}
\label{tab:uni_modal}
\begin{tabular}{@{}l|c|c|c|@{}}
\toprule
\textbf{Method} & \textbf{Error(\%)} & \textbf{Method} & \textbf{Error(\%)} \\
\midrule
ResNext-101 (baseline) & 90.0 & WSL & 54.9 \\
\midrule
$+$ TTA & 83.2 & $+$ TTA & 49.1 \\
\midrule
$+$ Single-point BN & 88.8 & $+$ Single-point BN & 58.9 \\
\midrule
$+$ MEMO & 84.3 & $+$ MEMO & 43.2 \\
\midrule
$+$ \textbf{ADTE} & \textbf{81.5} & $+$ \textbf{ADTE} & \textbf{41.1} \\
\bottomrule
\end{tabular}
\end{table}

Next, we conduct evaluations on five generalization models of CLIP. As shown in Table~\ref{tab:imagenet_five_models}, across all models, including SigLIP and SigLIP2, ADTE provides consistent improvements, confirming its robustness and wide applicability.

\begin{table}[h!]
\centering
\footnotesize
\caption{ImageNet-1k Accuracy (\%) on Generalization Models with ADTE.}
\label{tab:imagenet_five_models}
\vspace{-1em}
\begin{tabular}{@{}l rrr@{}}
\toprule
\textbf{Model} & \textbf{ImageNet-1k Acc. (\%)} & \textbf{$+$ADTE} \\
\midrule
CLIP & 68.7 & \textbf{71.8} \\
OpenCLIP & 70.2 & \textbf{73.5} \\
EVA-CLIP & 74.7 & \textbf{77.4} \\
SigLIP & 76.2 & \textbf{78.9} \\
SigLIP2 & 78.2 & \textbf{80.1} \\
\bottomrule
\end{tabular}
\end{table}

The above experiments confirm: (1) ADTE generalizes across architectures (unimodal or multimodal); (2) ADTE generalizes across training regimes (contrastive $\to$ sigmoid loss); (3) ADTE adapts robustly, even for models with stronger native calibration, such as SigLIP2.

\textbf{Confidence Intervals or Significance Tests..} To verify whether the improvements are statistically or practically meaningful, we add more runs to get the confidence intervals of accuracy. Table~\ref{tab:confidence_interval} demonstrates the statistical relevance of our improvements.

\begin{table}[h!]
\centering
\footnotesize
\caption{Accuracy (\%) 95\% confidence interval of the results in Table~\ref{tab:ablation study}.}
\vspace{-1em}
\label{tab:confidence_interval}
\begin{tabular}{@{}l rrrrr@{}}
\toprule
\textbf{Model} & \textbf{ImageNet-1k} & \textbf{ImageNet-A} & \textbf{ImageNet-V} & \textbf{ImageNet-K} & \textbf{ImageNet-R} \\
\midrule
CLIP-ViT/B-16 & 68.7 & 50.6 & 62.2 & 48.3 & 77.7 \\
SE & 70.9 & 64.0 & 65.1 & 50.3 & 80.8 \\
ADTE & 71.8 & 65.5 & 65.6 & 53.5 & 81.4 \\
ADTE\_95\% con. int. & $71.78\pm0.44$ & $65.72\pm0.50$ & $65.67\pm0.17$ & $53.59\pm0.25$ & $81.28\pm0.25$ \\
\bottomrule
\end{tabular}
\end{table}

\section{Proof}
\subsection{Proof of property~\ref{property-1}}\label{Proof of property 1}
$\blacktriangleright$ \texttt{Shannon-Tsallis $q \to 1$ Equivalence:}

\begin{property}
As $q \to 1$, Tsallis Entropy becomes equivalent to Shannon Entropy, i.e.,
\begin{footnotesize}
\begin{equation}
    \lim_{q\to 1}\mathbf{H}_{\textup{\texttt{TE}}}(\mathbf{P}(\cdot\mid\mathbf{x}_j^{\textup{test}})) = \mathbf{H}_\textup{\texttt{SE}}(\mathbf{P}(\cdot\mid\mathbf{x}_j^{\textup{test}})).
\end{equation}
\end{footnotesize}
\end{property}
\proof{
For Tsallis entropy, we have:
\begin{footnotesize}
\begin{equation}
\begin{split}
    \mathbf{H}_\texttt{TE}(\mathbf{P}(\cdot\mid\mathbf{x}_j^{\text{test}})) =& \frac{1}{1-q}\left( \sum_{l=1}^L \mathbf{P}(y=l\mid\mathbf{x}_j^{\text{test}})^q - 1 \right) = \frac{1}{1-q}\left( \sum_{l=1}^L \mathbf{P}(y=l\mid\mathbf{x}_j^{\text{test}})^q - \sum_{l=1}^L \mathbf{P}(y=l\mid\mathbf{x}_j^{\text{test}}) \right) \\
    =& \sum_{l=1}^L \mathbf{P}(y=l\mid\mathbf{x}_j^{\text{test}}) \left( \frac{\mathbf{P}(y=l\mid\mathbf{x}_j^{\text{test}})^{q-1}-1}{1-q} \right).
\end{split}    
\end{equation}
\end{footnotesize}%
The limitation $\lim_{q\to1}$ only applies to the terms containing $q$. Thus:
\begin{footnotesize}
\begin{equation}
\begin{split}
    \lim_{q\to1}\mathbf{H}_\texttt{TE}(\mathbf{P}(\cdot\mid\mathbf{x}_j^{\text{test}})) =& \sum_{l=1}^L \mathbf{P}(y=l\mid\mathbf{x}_j^{\text{test}}) \left( \lim_{q\to1} \frac{\mathbf{P}(y=l\mid\mathbf{x}_j^{\text{test}})^{q-1}-1}{1-q} \right).
\end{split}    
\label{equ:TE_Limitation_appendix}
\end{equation}
\end{footnotesize}%
We use L'Hopital's rule to solve the limit in the expression:
\begin{footnotesize}
\begin{equation}
\begin{split}
    \lim_{q\to1} \frac{\mathbf{P}(y=l\mid\mathbf{x}_j^{\text{test}})^{q-1}-1}{1-q} =&
    \lim_{\gamma\to0} \frac{\mathbf{P}(y=l\mid\mathbf{x}_j^{\text{test}})^{\gamma}-1}{-\gamma} = \lim_{\gamma\to0} \frac{\mathbf{P}(y=l\mid\mathbf{x}_j^{\text{test}})^{\gamma} \log \mathbf{P}(y=l\mid\mathbf{x}_j^{\text{test}})}{-1} \\
    =& \frac{1 \times \log \mathbf{P}(y=l\mid\mathbf{x}_j^{\text{test}})}{-1} = - \log \mathbf{P}(y=l\mid\mathbf{x}_j^{\text{test}}),
\end{split}    
\end{equation}
\end{footnotesize}%
where the second equality uses L'Hopital's rule.  Substituting the above into Equation~\ref{equ:TE_Limitation_appendix}, we have:
\begin{footnotesize}
\begin{equation}
\begin{split}
    \lim_{q\to1}\mathbf{H}_\texttt{TE}(\mathbf{P}(\cdot\mid\mathbf{x}_j^{\text{test}})) =& \sum_{l=1}^L \mathbf{P}(y=l\mid\mathbf{x}_j^{\text{test}}) \left( - \log \mathbf{P}(y=l\mid\mathbf{x}_j^{\text{test}}) \right) \\
    =& -\sum_{l=1}^{L} \mathbf{P}(y = l\mid\mathbf{x}_j^{\text{test}}) \log \mathbf{P}(y = l\mid\mathbf{x}_j^{\text{test}})\\
    =& \mathbf{H}_\texttt{SE}(\mathbf{P}(\cdot\mid\mathbf{x}_j^{\text{test}})).
\end{split}    
\end{equation}
\end{footnotesize}
Therefore, we prove Property~\ref{property-1}.

\begin{figure}
    \centering
    \rotatebox{270}{\includegraphics[width=1.4\linewidth]{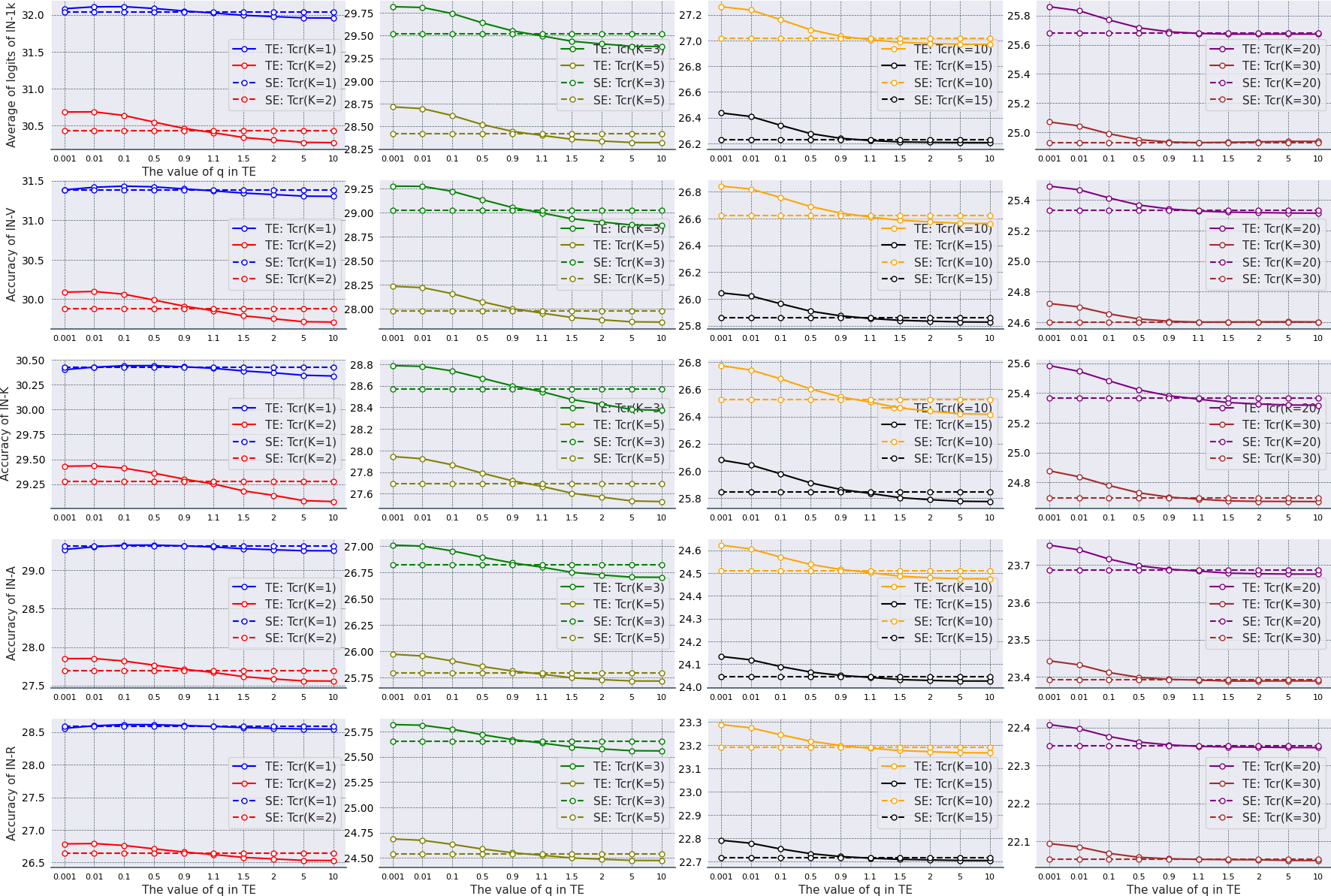}} 
    \captionsetup{font={footnotesize}}
    \caption{Average $\mathbf{Tcr}_K$ values for different $q$ of TE and SE on ImageNet-1K and its variant datasets.}
    \label{fig:average_logits}
\end{figure}

\subsection{Experimental analysis of property~\ref{property-2}}\label{proof of property 2}
$\blacktriangleright$ \texttt{Higher $\mathbf{Tcr}_K$ under TE as $q\searrow$ and Comparison with SE:}
\begin{property}\label{property-2}
Through experimental analysis, we find that as the parameter $q$ decreases, the set of high-confidence views selected by \texttt{TE} tends to have a higher average $\mathbf{Tcr}_K$ value (for $K > 1$). For two different parameter values $q_1$ and $q_2$ with $q_1 < q_2$, and their corresponding selected view sets $\mathcal{X}_{q_1}$ and  $\mathcal{X}_{q_2}$ of equal size, we observe:
\begin{equation}\label{eq:tcr_te}
    \frac{1}{|\mathcal{X}_{q_1}|}\sum_{\mathbf{x_1}\in \mathcal{X}_{q_1}} \mathbf{Tcr}_{K}(\mathbf{x_1}) > \frac{1}{|\mathcal{X}_{q_2}|} \sum_{\mathbf{x_2}\in \mathcal{X}_{q_2}} \mathbf{Tcr}_{K}(\mathbf{x_2}).
\end{equation}
Furthermore, there exists a particular $q^*$ and corresponding view set $\mathcal{X}_{q^*}^{\texttt{\textup{TE}}}$, \texttt{TE} outperforms \texttt{SE} in this regard, with the view set selected by \texttt{SE} denoted as $\mathcal{X}^{\texttt{\textup{SE}}}$:
\begin{equation}
     \frac{1}{|\mathcal{X}_{q^*}^{\texttt{\textup{TE}}}|} \sum_{\mathbf{x_1}\in \mathcal{X}_{q^*}^{\texttt{\textup{TE}}}} \mathbf{Tcr}_{K}(\mathbf{x_1}) > \frac{1}{|\mathcal{X}^{\texttt{\textup{SE}}}|} \sum_{\mathbf{x_2}\in \mathcal{X}^{\texttt{\textup{SE}}}} \mathbf{Tcr}_{K}(\mathbf{x_2}).
\end{equation}\label{tcr_te_se}
\textit{Experimental analysis.} \textup{Equation~\ref{eq:tcr_te} describes that when $K > 1$, as the parameter $q$ of \texttt{TE} decreases, the selected confidence views exhibit higher average $\mathbf{Tcr}_K$ values. For example, in Figure~\ref{fig:average_logits} of ImageNet-1k and its five variants, as the $q$ value decreases from $10$ to $0.001$, except for $\mathbf{Tcr}(K=1)$, all other $\mathbf{Tcr}$ values show a gradually increasing trend. This implies that the lower the $q$ value, the more the $TE$ tends to select views with higher similarity scores.}

\textup{As shown in Figure~\ref{fig:average_logits}, when the $q$ value of \texttt{TE} is relatively large, the $\mathbf{Tcr}$ value of \texttt{TE} is generally smaller than that of \texttt{SE}. As the $q$ value decreases, the $\mathbf{Tcr}$ value of \texttt{TE} gradually exceeds that of \texttt{SE}. This further indicates that \texttt{SE} is a special case of \texttt{TE}, and appropriate selection of $q$ values can make \texttt{TE} exhibit better performance than \texttt{SE}.}
\end{property}


\subsection{Proof of conclusions~\ref{Correcting Biased Entropy}}\label{proof of Correcting Biased Entropy}
Let $\mathbf{p} \in (0,\epsilon)$, where $\epsilon$ ensures the probability of $\mathbf{p}$ of \texttt{tail} category being close to zero, and we have the following conclusions:

\vspace{1em}
$\blacktriangleright$ \underline{\texttt{(1) As $q\!\to \infty,$ we have $\lim_{\mathbf{p}\to0^+}\mathbf{F}(\mathbf{p},q)=0^-,$ and $\mathbf{F}(\mathbf{p}, q_1)\!<\!\mathbf{F}(\mathbf{p}, q_2) < 0,$ $\forall q_1\!<\!q_2\!<\infty$.}}
\proof{
    
    
    a. Find the limit of the function $\mathbf{F}(\mathbf{p},q) = \frac{\mathbf{p}^q}{1-q} + \mathbf{p}\log \mathbf{p}$ as $q \to +\infty$ and $p \to 0^+$.
    
    \texttt{First term}: $\frac{\mathbf{p}^q}{1-q} \to 0^-$ as $q \to +\infty$ and $\mathbf{p} \to 0^+$.
    
    \texttt{Second term}: $\mathbf{p}\log \mathbf{p} \to 0^-$ by L'Hospital's rule: $\lim_{\mathbf{p} \to 0^+} \frac{\log \mathbf{p}}{1/\mathbf{p}} = \lim_{\mathbf{p} \to 0^+} (-\mathbf{p}) = 0^-$
    
    \texttt{Conclusion}:
    \begin{equation}
    \lim_{\substack{q \to +\infty \\ \mathbf{p} \to 0^+}} \mathbf{F}(\mathbf{p},q) = 0^- + 0^- = 0^-
    \end{equation}

    b. Compare the magnitudes of $\mathbf{F}(\mathbf{p},q_1)$, $\mathbf{F}(\mathbf{p},q_2)$, and 0 as $q_1 < q_2 \to +\infty$ and $\mathbf{p} \to 0^+$.

    \texttt{Sign analysis}:
    For $\mathbf{p} \in (0,\epsilon)$ and $q > 1$:
    \begin{equation}
    \mathbf{F}(\mathbf{p},q) = \underbrace{\frac{\mathbf{p}^q}{1-q}}_{<0} + \underbrace{\mathbf{p}\log \mathbf{p}}_{<0} < 0
    \end{equation}
    
    \texttt{Monotonicity}:
    Fix $\mathbf{p} \in (0,\epsilon)$. The derivative of $\mathbf{G}(q) = \frac{\mathbf{p}^q}{1-q}$:
    \begin{equation}
    \mathbf{G}'(q) = \frac{\mathbf{p}^q [1+(1-q)\log \mathbf{p}]}{(1-q)^2} > 0 \quad \text{for } q \to +\infty
    \end{equation}
    Thus $\mathbf{G}(q)$ is increasing. For $q_1 < q_2$:
    \begin{equation}
    \mathbf{G}(q_1) < \mathbf{G}(q_2) \implies \mathbf{F}(\mathbf{p},q_1) < \mathbf{F}(\mathbf{p},q_2) < 0
    \end{equation}
}

\vspace{1em}
$\blacktriangleright$ \underline{\texttt{(2) As $q\!\to\!1^{+},$ we have $\lim_{\mathbf{p}\to0^+}\mathbf{F}(\mathbf{p},q)\!=\!-\infty,$ and $\mathbf{F}(\mathbf{p}, q_1)\!<\!\mathbf{F}(\mathbf{p}, q_2)\!<\!0,$ $\forall 1\!<\!q_1\!<\!q_2$.}}
\proof{
    a. Find the limit of the function $\mathbf{F}(\mathbf{p},q) = \frac{\mathbf{p}^q}{1-q} + \mathbf{p}\log \mathbf{p}$ as $q \to 1^+$ and $p \to 0^+$.
    
    \texttt{Consider the limit of the entire function} $\mathbf{F}(\mathbf{p},q)$: Since there are two variables approaching their limits, the path of the limit may affect the result. We consider the iterated limits:
    
    \ding{172} \texttt{First, fix $q > 1$ and let $\mathbf{p} \to 0^+$}.
    At this time, we have the $\lim_{\mathbf{p}\to 0^+} \mathbf{F}(\mathbf{p},q) = \lim_{\mathbf{p}\to 0^+} \left( \frac{\mathbf{p}^q}{1-q} + \mathbf{p}\log \mathbf{p} \right)$.
    Since $q > 1$ is fixed, $\lim_{\mathbf{p}\to 0^+} \frac{\mathbf{p}^q}{1-q} = \frac{0^q}{1-q} = 0^-$, and we have already found that $\lim_{\mathbf{p}\to 0^+} \mathbf{p}\log \mathbf{p} = 0^-$. So, $\lim_{\mathbf{p}\to 0^+} \mathbf{F}(\mathbf{p},q) = 0^- + 0^- = 0^-$.
    Then, let $q \to 1^+$, then $\lim_{q\to 1^+} (0^-) = 0^-$.
    
    \ding{173} \texttt{Next, fix $\mathbf{p} \in (0,\epsilon)$ and let $q \to 1^+$}.
    For $\lim_{q\to 1^+} \mathbf{F}(\mathbf{p},q) = \lim_{q\to 1^+} \left( \frac{\mathbf{p}^q}{1-q} + \mathbf{p}\log \mathbf{p} \right)$.
    When $q \to 1^+$, $1 - q \to 0^-$, $\mathbf{p}^q \to \mathbf{p}$, so $\lim_{q\to 1^+} \frac{\mathbf{p}^q}{1-q} = \frac{\mathbf{p}}{0^-} = -\infty$ (because $\mathbf{p} > 0$), and for a fixed $\mathbf{p} \in (0,\epsilon)$, $\mathbf{p}\log \mathbf{p}$ is a fixed negative value. So, $\lim_{q\to 1^+} \mathbf{F}(\mathbf{p},q) = -\infty + \mathbf{p}\log \mathbf{p} = -\infty$.
    Then, let $\mathbf{p} \to 0^+$, then $\lim_{\mathbf{p}\to 0^+} (-\infty) = -\infty$.

    Since the results of the two iterative limits are different (one is $0$ and the other is $-\infty$), strictly speaking, the existence of the simultaneous limit depends on the relative rates at which $\mathbf{p}$ and $ q-1$ approach $0$. However, in this work, the probability of the \texttt{tail} category is usually a very small decimal, but it will not approach $0^+$ infinitely, and we give priority to the case of $q \to 1^+$. Therefore, $\lim_{q\to 1^+} \mathbf{F}(\mathbf{p},q) = -\infty + \mathbf{p}\log \mathbf{p} = -\infty$. Therefore, we determine the limit value to be $-\infty$.

    b. Compare the magnitudes of $\mathbf{F}(\mathbf{p},q_1)$, $\mathbf{F}(\mathbf{p},q_2)$, and 0 as $q_1 < q_2 \to 1^+$ and $\mathbf{p} \to 0^+$.

    \texttt{This part of the proof is the same as part b of Conclusion~\texttt{(1)}}.
    
}

$\blacktriangleright$ \underline{\texttt{(3) As $q\!\to\!0^{+},$ we have $\lim_{\mathbf{p}\to0^{+}}\mathbf{F}(\mathbf{p},q)\!=\!1^-,$ and $\mathbf{F}(\mathbf{p}, q_1)\!>\!\mathbf{F}(\mathbf{p}, q_2)\!>\!0,$ $\forall 0\!<\!q_1\!<\!q_2$.}}
\proof{
    a. Find the limit of the function $\mathbf{F}(\mathbf{p},q) = \frac{\mathbf{p}^q}{1-q} + \mathbf{p}\log \mathbf{p}$ as $q \to 1^+$ and $p \to 0^+$.

    In this condition, both $\mathbf{p}$ and $q$ approach $0$ simultaneously. This is a limit problem for a two-variable function. We need to examine path dependence.
    
    \texttt{Consider different paths $(\mathbf{p},q) \to (0^+,0^+)$}:
    
    \ding{172} \texttt{First: Iterated limit with \( q \to 0^+ \) first, then \( \mathbf{p} \to 0^+ \).}
    \begin{equation}
    \lim_{q \to 0^+} \mathbf{F}(\mathbf{p},q) = \lim_{q \to 0^+} \left[ \frac{\mathbf{p}^q}{1-q} + \mathbf{p} \log \mathbf{p} \right] = 1^- + \mathbf{p} \log \mathbf{p}.
    \end{equation}
    Thus,
    \begin{equation}
    \lim_{\mathbf{p} \to 0^+} \left(1^- + \mathbf{p} \log \mathbf{p}\right) = 1^-.
    \end{equation}
    
    \ding{173} \texttt{Next: Iterated limit with \( \mathbf{p} \to 0^+ \) first, then \( q \to 0^+ \).} For fixed \( q \in (0,1) \):
    \begin{equation}
    \lim_{\mathbf{p} \to 0^+} \mathbf{F}(\mathbf{p},q) = \lim_{\mathbf{p} \to 0^+} \left[ \frac{\mathbf{p}^q}{1-q} + \mathbf{p} \log \mathbf{p} \right] = 0.
    \end{equation}
    Therefore,
    \begin{equation}
    \lim_{q \to 0^+} \left(0\right) = 0.
    \end{equation}


    In this work, $\mathbf{p}$ represents the probability of a category, which may be a very small decimal, but it will not approach $0$ infinitely. \texttt{Path 2} can be ignored. Thus, we determine the limit value to be $1^-$.

    b. Compare the magnitudes of $\mathbf{F}(\mathbf{p},q_1)$, $\mathbf{F}(\mathbf{p},q_2)$, and 0 as $q_1 < q_2 \to 0^+$ and $\mathbf{p} \to 0^+$.

    
    \texttt{Sign Analysis}: As \(\mathbf{p} \to 0^+\), \(\mathbf{p} \log \mathbf{p} \to 0^-\). For positivity:
    \begin{equation}
    \mathbf{F}(\mathbf{p},q) > 0 \iff \frac{\mathbf{p}^q}{1-q} > -\mathbf{p} \log \mathbf{p} \iff \frac{\mathbf{p}^{q-1}}{1-q} > |\log \mathbf{p}|.
    \end{equation}
    As \(q \to 0^+\), \(q-1 \approx -1\) and \(1-q \approx 1\). The inequality simplifies to: $\frac{1}{\mathbf{p}} > |\log \mathbf{p}|$. Since \(\frac{1}{\mathbf{p}}\) diverges faster than \(|\log \mathbf{p}|\) as \(\mathbf{p} \to 0^+\), \(\mathbf{F}(\mathbf{p},q) > 0\) for sufficiently small \(\mathbf{p}\). Thus:
    \begin{equation}
    \mathbf{F}(\mathbf{p}, q_1) > 0 \quad \text{and} \quad \mathbf{F}(\mathbf{p}, q_2) > 0.
    \end{equation}
    
    \texttt{Monotonicity in \(q\)}: Fix \(\mathbf{p}\). The derivative w.r.t. \(q\) is:
    \begin{equation}
    \mathbf{G}'(q) = \frac{\mathbf{p}^q [1 + (1-q)\log \mathbf{p}]}{(1-q)^2}.
    \end{equation}
    For \(\mathbf{p} \to 0^+\), \(\log \mathbf{p} \to -\infty\). Since \(1-q \approx 1\), the numerator is negative, implying \(\mathbf{G}'(q) < 0\). Thus, \(\mathbf{F}(\mathbf{p},q)\) is decreasing in \(q\). For \(q_1 < q_2\):
    \begin{equation}
    0 < \mathbf{F}(\mathbf{p}, q_2) < \mathbf{F}(\mathbf{p}, q_1).
    \end{equation}
}

$\blacktriangleright$ \underline{\texttt{(4) As $q\!\to\!1^-,$ we have $\lim_{\mathbf{p}\to0^+}\mathbf{F}(\mathbf{p},q)=+\infty,$ and $\mathbf{F}(\mathbf{p}, q_2)\!>\!\mathbf{F}(\mathbf{p}, q_1)\!>\!0,$ $\forall q_1\!<\!q_2\!<1$.}}
\proof{

    a. Find the limit of the function $\mathbf{F}(\mathbf{p},q) = \frac{\mathbf{p}^q}{1-q} + \mathbf{p}\log \mathbf{p}$ as $q \to 1^-$ and $p \to 0^+$.

    
    
    \texttt{Compute the iterated limits}:
    
    \ding{172} For fixed \( q \in (0,1) \):
    \begin{equation}
    \lim_{\mathbf{p} \to 0^+} \mathbf{F}(\mathbf{p}, q) = \lim_{\mathbf{p} \to 0^+} \left[ \frac{\mathbf{p}^q}{1-q} + \mathbf{p} \log \mathbf{p} \right] = 0^+ + 0^- = 0^+.
    \end{equation}
    Thus,
    \begin{equation}
    \lim_{q \to 1^-} \left( 0^+ \right) = 0^+.
    \end{equation}
    
    \ding{173} For fixed \( \mathbf{p} \in (0, \epsilon) \):
    \begin{equation}
    \lim_{q \to 1^-} \mathbf{F}(\mathbf{p}, q) = \lim_{q \to 1^-} \left[ \frac{\mathbf{p}^q}{1-q} + \mathbf{p} \log \mathbf{p} \right] = +\infty + \mathbf{p} \log \mathbf{p} = +\infty.
    \end{equation}
    In this work, the probability of the \texttt{tail} category is usually a very small decimal, but it will not approach $0^+$ infinitely, and we give priority to the case of $\mathbf{q} \to 1^-$. Therefore, $\lim_{q\to 1^-} \mathbf{F}(\mathbf{p},q) = +\infty$. Therefore, we determine the limit value to be $+\infty$.
    
    b. Compare the magnitudes of $\mathbf{F}(\mathbf{p},q_1)$, $\mathbf{F}(\mathbf{p},q_2)$, and 0 as $q_1 < q_2 \to 1^-$ and $\mathbf{p} \to 0^+$.

    \texttt{Sign analysis}: Identical to part sign analysis of Conclusion (3).

    \texttt{Comparing $\mathbf{F}(\mathbf{p}, q_1)$ and $\mathbf{F}(\mathbf{p}, q_2)$}: Compute the partial derivative:
    \begin{equation}
    \frac{\partial \mathbf{F}}{\partial q} = \frac{\mathbf{p}^q}{1-q} \left[ \log \mathbf{p} + \frac{1}{1-q} \right].
    \end{equation}
    Since $\frac{\mathbf{p}^q}{1-q} > 0$, the sign of $\frac{\partial \mathbf{F}}{\partial q}$ depends on $\left[ \log \mathbf{p} + \frac{1}{1-q} \right]$. As $q \to 1^-$, $\frac{1}{1-q} \to +\infty$ dominates $\log \mathbf{p} \to -\infty$. For example:
    \[
    \mathbf{p} = 0.01 \implies \log \mathbf{p} \approx -4.6; \quad q = 0.999 \implies \frac{1}{1-q} = 1000 \implies \log \mathbf{p} + \frac{1}{1-q} > 0.
    \]
    Thus, $\mathbf{F}(\mathbf{p}, q)$ increases with $q$ near $q=1$. For $q_1 < q_2 \to 1^-$, we have:
    \begin{equation}
    0 < \mathbf{F}(\mathbf{p}, q_1) < \mathbf{F}(\mathbf{p}, q_2).
    \end{equation}
}

\end{document}

%% file: math_commands.tex
\usepackage{amsmath,amsfonts,bm}

\def\eqref#1{equation~\ref{#1}}

\def\1{\bm{1}}

\DeclareMathAlphabet{\mathsfit}{\encodingdefault}{\sfdefault}{m}{sl}
\SetMathAlphabet{\mathsfit}{bold}{\encodingdefault}{\sfdefault}{bx}{n}



%% file: algos.tex


\begin{figure}
    \centering
    \begin{minipage}{0.45\linewidth}
        \begin{algorithm}[H]
            \caption{Estimating the bias $\mathbf{b}$}
            \begin{algorithmic}[1]
                \State \textbf{Input}: Estimated  $\mathbb{E}_{\mathrm{x}\sim \mathbf{P}(\mathbf{x}|l^{'})} [ \mathbf{P}(l|\mathbf{x})]$, maximum number of iteration $T$, convergence threshold $\varepsilon$
                \State \textbf{Output}: Estimated $\mathrm{b}$
                \State $a_{ll'} := \mathbb{E}_{\mathrm{x}\sim \mathbf{P}(\mathbf{x}|l^{'})} [ \mathbf{P}(l|\mathbf{x})]$
                \State $\tilde{\mathrm{p}}^{(0)} \gets [\frac{1}{|\mathcal{Y}^{\text{test}}|},\dots, \frac{1}{|\mathcal{Y}^{\text{test}}|}]$
                \For{$t$: $0, \dots, T-1$}
                    \State $\tilde{\mathrm{p}}_l^{(t + 1)} \leftarrow \sum_{l' \in \mathcal{Y}^{\text{test}}} \tilde{\mathrm{p}}_{l'}^{(t)} \cdot a_{ll'}$
                    \State $\tilde{\mathrm{p}}^{(t+1)} \gets \tilde{\mathrm{p}}^{(t+1)} / \| \tilde{\mathrm{p}}^{(t+1)} \|_1$
                    \If{$\|\tilde{\mathrm{p}}^{(t+1)}-\tilde{\mathrm{p}}^{(t)}\|_1 \leq \varepsilon$}
                        \State \textbf{break}
                    \EndIf
                \EndFor
                \State \textbf{return} $-\log{\tilde{\mathrm{p}}^{(t)}}$
            \end{algorithmic}\label{algo:estimate_b}
        \end{algorithm}
    \end{minipage}
    \hfill
    \begin{minipage}{0.53\linewidth}
        \begin{algorithm}[H]
            \begin{spacing}{0.94}
            \caption{Pipeline of \textbf{A}daptive \textbf{D}ebiasing \textbf{T}sallis \textbf{E}ntropy~(\texttt{ADTE}) for Test-Time Adaptation~(TTA)}
            \begin{algorithmic}[1]
                \State \textbf{Input}: Estimated bias $\mathrm{b}$, test input image $\mathbf{x}^{\text{test}}$, the number of image augmentation $N$, the number of selected confident views $N_v$
                \State \textbf{Output}: Prediction $\hat{y}$
                \State views set $\mathbf{X}^{\text{test}}$ $\gets$ augment($\mathbf{x}^{\text{test}}$, num\_views=N)
                \State Calculate $\mathbf{P}(\cdot | \mathbf{x}_j^{\text{test}}), \forall \mathbf{x}_j^{\text{test}}\in \mathbf{X}^{\text{test}}$
                \State Calculate $q^l$ for each class with Equation~\ref{equ:calcu_q}
                \State Calculate $\mathbf{H}_\texttt{ADTE}(\mathbf{P}(\cdot | \mathbf{x}_j^{\text{test}})), \forall \mathbf{x}_j^{\text{test}}\in \mathbf{X}^{\text{test}}$ with Equation~\ref{equ:ADTE}
                \State Confident views set $\mathbf{\hat{X}}^{\text{test}}$ $\gets$ top $N_v$ views with the smallest $\mathbf{H}_\texttt{ADTE}$ values
                \State $\tilde{\mathbf{P}} \gets$aggregate($\{\mathbf{P}(\cdot | \mathbf{\hat{x}}_j^{\text{test}}) |$ $\forall \mathbf{\hat{x}}_j^{\text{test}} \in \mathbf{\hat{X}}^{\text{test}} \}$)
                \State $\hat{y} \gets \arg\max_l \tilde{\mathbf{P}}(y=l | \cdot)$
                \State \textbf{return} $\hat{y}$
            \end{algorithmic}\label{algo:pipeline}
            \end{spacing}
        \end{algorithm}
    \end{minipage}
\end{figure}

%% file: iclr2026_conference.bib
@article{VLMs-XVLM2,
  author       = {Yan Zeng and
                  Xinsong Zhang and
                  Hang Li and
                  Jiawei Wang and
                  Jipeng Zhang and
                  Wangchunshu Zhou},
  title        = {X2-VLM: All-in-One Pre-Trained Model for Vision-Language Tasks},
  journal      = {TPAMI},
  volume       = {46},
  number       = {5},
  pages        = {3156--3168},
  year         = {2024},
}

@inproceedings{VLMs-Openai-CLIP,
  author       = {Alec Radford and
                  Jong Wook Kim and
                  Chris Hallacy and
                  Aditya Ramesh and
                  Gabriel Goh and
                  Sandhini Agarwal and
                  Girish Sastry and
                  Amanda Askell and
                  Pamela Mishkin and
                  Jack Clark and
                  Gretchen Krueger and
                  Ilya Sutskever},
  title        = {Learning Transferable Visual Models From Natural Language Supervision},
  booktitle    = {ICML},
  volume       = {139},
  pages        = {8748--8763},
  year         = {2021},
}

@inproceedings{ITP-CC3M,
  author       = {Piyush Sharma and
                  Nan Ding and
                  Sebastian Goodman and
                  Radu Soricut},
  title        = {Conceptual Captions: {A} Cleaned, Hypernymed, Image Alt-text Dataset For Automatic Image Captioning},
  booktitle    = {ACL},
  pages        = {2556--2565},
  year         = {2018},
}

@inproceedings{ITP-Laion-5b,
  title={Laion-5b: An open large-scale dataset for training next generation image-text models},
  author={Schuhmann, Christoph and Beaumont, Romain and Vencu, Richard and Gordon, Cade and Wightman, Ross and Cherti, Mehdi and Coombes, Theo and Katta, Aarush and Mullis, Clayton and Wortsman, Mitchell and others},
  booktitle={NeurIPS},
  volume={35},
  pages={25278--25294},
  year={2022}
}

@article{Prompt-CoOp,
  author       = {Kaiyang Zhou and
                  Jingkang Yang and
                  Chen Change Loy and
                  Ziwei Liu},
  title        = {Learning to Prompt for Vision-Language Models},
  journal      = {IJCV},
  volume       = {130},
  number       = {9},
  pages        = {2337--2348},
  year         = {2022},
}

@article{Prompt-DualCoOp++,
  title={Dualcoop++: Fast and effective adaptation to multi-label recognition with limited annotations},
  author={Hu, Ping and Sun, Ximeng and Sclaroff, Stan and Saenko, Kate},
  journal={TPAMI},
  year={2023},
  publisher={IEEE}
}

@inproceedings{Prompt-VLPL,
  title={Vision-Language Pseudo-Labels for Single-Positive Multi-Label Learning},
  author={Xing, Xin and Xiong, Zhexiao and Stylianou, Abby and Sastry, Srikumar and Gong, Liyu and Jacobs, Nathan},
  booktitle={CVPR},
  pages={7799--7808},
  year={2024}
}

@inproceedings{TTA-Tent,
  author       = {Dequan Wang and
                  Evan Shelhamer and
                  Shaoteng Liu and
                  Bruno A. Olshausen and
                  Trevor Darrell},
  title        = {Tent: Fully Test-Time Adaptation by Entropy Minimization},
  booktitle    = {ICLR},
  year         = {2021},
}

@inproceedings{TTA-Delta,
  author       = {Bowen Zhao and
                  Chen Chen and
                  Shu{-}Tao Xia},
  title        = {Delta: Degradation-Free Fully Test-Time Adaptation},
  booktitle    = {ICLR},
  year         = {2023},
}

@inproceedings{TTA-Stationary,
  author       = {Jae{-}Hong Lee and
                  Joon{-}Hyuk Chang},
  title        = {Stationary Latent Weight Inference for Unreliable Observations from Online Test-Time Adaptation},
  booktitle    = {ICML},
  year         = {2024},
}

@inproceedings{TTA-NotEnough,
  author       = {Jonghyun Lee and
                  Dahuin Jung and
                  Saehyung Lee and
                  Junsung Park and
                  Juhyeon Shin and
                  Uiwon Hwang and
                  Sungroh Yoon},
  title        = {Entropy is not Enough for Test-Time Adaptation: From the Perspective of Disentangled Factors},
  booktitle    = {ICLR},
  year         = {2024},
}

@inproceedings{TTA-VIDA,
  author       = {Jiaming Liu and
                  Senqiao Yang and
                  Peidong Jia and
                  Renrui Zhang and
                  Ming Lu and
                  Yandong Guo and
                  Wei Xue and
                  Shanghang Zhang},
  title        = {ViDA: Homeostatic Visual Domain Adapter for Continual Test Time Adaptation},
  booktitle    = {ICLR},
  year         = {2024},
}

@inproceedings{TTA-EcoTTA,
  author       = {Junha Song and
                  Jungsoo Lee and
                  In So Kweon and
                  Sungha Choi},
  title        = {EcoTTA: Memory-Efficient Continual Test-Time Adaptation via Self-Distilled Regularization},
  booktitle    = {CVPR},
  pages        = {11920--11929},
  year         = {2023},
}

@inproceedings{TTA-TPT,
  author       = {Manli Shu and
                  Weili Nie and
                  De{-}An Huang and
                  Zhiding Yu and
                  Tom Goldstein and
                  Anima Anandkumar and
                  Chaowei Xiao},
  title        = {Test-Time Prompt Tuning for Zero-Shot Generalization in Vision-Language Models},
  booktitle    = {NeurIPS},
  year         = {2022},
}

@inproceedings{TTA-MEMO,
  author       = {Marvin Zhang and
                  Sergey Levine and
                  Chelsea Finn},
  title        = {MEMO: Test Time Robustness via Adaptation and Augmentation},
  booktitle    = {NeurIPS},
  year         = {2022},
}

@inproceedings{TTA-DiffTPT,
  author       = {Chun{-}Mei Feng and
                  Kai Yu and
                  Yong Liu and
                  Salman Khan and
                  Wangmeng Zuo},
  title        = {Diverse Data Augmentation with Diffusions for Effective Test-time Prompt Tuning},
  booktitle    = {ICCV},
  pages        = {2704--2714},
  year         = {2023},
}

@inproceedings{TTA-TDA,
  title={Efficient Test-Time Adaptation of Vision-Language Models},
  author={Karmanov, Adilbek and Guan, Dayan and Lu, Shijian and El Saddik, Abdulmotaleb and Xing, Eric},
  booktitle={CVPR},
  pages={14162--14171},
  year={2024}
}

@inproceedings{TTA-Zero,
  author       = {Matteo Farina and
                  Gianni Franchi and
                  Giovanni Iacca and
                  Massimiliano Mancini and
                  Elisa Ricci},
  title        = {Frustratingly Easy Test-Time Adaptation of Vision-Language Models},
  booktitle    = {NeurIPS},
  year         = {2024},
}

@inproceedings{TTA-ML-TTA,
  author       = {Xiangyu Wu and
                  Feng Yu and
                  Qing{-}Guo Chen and
                  Yang Yang and
                  Jianfeng Lu},
  title        = {Multi-Label Test-Time Adaptation with Bound Entropy Minimization},
  booktitle    = {ICLR},
  year         = {2025},
}

@inproceedings{TTA-DynaPrompt,
  author       = {Zehao Xiao and
                  Shilin Yan and
                  Jack Hong and
                  Jiayin Cai and
                  Xiaolong Jiang and
                  Yao Hu and
                  Jiayi Shen and
                  Qi Wang and
                  Cees G. M. Snoek},
  title        = {DynaPrompt: Dynamic Test-Time Prompt Tuning},
  booktitle    = {ICLR},
  year         = {2025},
}

@inproceedings{TTA-Frolic,
  author       = {Xingyu Zhu and
                  Beier Zhu and
                  Yi Tan and
                  Shuo Wang and
                  Yanbin Hao and
                  Hanwang Zhang},
  title        = {Enhancing Zero-Shot Vision Models by Label-Free Prompt Distribution
                  Learning and Bias Correcting},
  booktitle    = {NeurIPS},
  year         = {2024},
}

@inproceedings{TTA-BCA,
  author       = {Lihua Zhou and
                  Mao Ye and
                  Shuaifeng Li and
                  Nianxin Li and
                  Xiatian Zhu and
                  Lei Deng and
                  Hongbin Liu and
                  Zhen Lei},
  title        = {Bayesian Test-Time Adaptation for Vision-Language Models},
  booktitle    = {CVPR},
  year         = {2025},
}

@inproceedings{EM-01,
  title={Semi-supervised learning by entropy minimization},
  author={Grandvalet, Yves and Bengio, Yoshua},
  booktitle={NeurIPS},
  volume={17},
  year={2004}
}

@inproceedings{EM-02,
  author       = {David Berthelot and
                  Nicholas Carlini and
                  Ian J. Goodfellow and
                  Nicolas Papernot and
                  Avital Oliver and
                  Colin Raffel},
  title        = {MixMatch: A Holistic Approach to Semi-Supervised Learning},
  booktitle    = {NeurIPS},
  pages        = {5050--5060},
  year         = {2019},
}

@inproceedings{EM-03,
  author       = {Kuniaki Saito and
                  Donghyun Kim and
                  Stan Sclaroff and
                  Trevor Darrell and
                  Kate Saenko},
  title        = {Semi-Supervised Domain Adaptation via Minimax Entropy},
  booktitle    = {ICCV},
  pages        = {8049--8057},
  year         = {2019},
}

@article{EM-04,
  author       = {Obsa Gilo and
                  Jimson Mathew and
                  Samrat Mondal and
                  Rakesh Kumar Sandoniya},
  title        = {Subdomain adaptation via correlation alignment with entropy minimization for unsupervised domain adaptation},
  journal      = {Pattern Anal. Appl.},
  volume       = {27},
  number       = {1},
  pages        = {13},
  year         = {2024},
}

@inproceedings{LA-01,
  author       = {Aditya Krishna Menon and
                  Sadeep Jayasumana and
                  Ankit Singh Rawat and
                  Himanshu Jain and
                  Andreas Veit and
                  Sanjiv Kumar},
  title        = {Long-tail learning via logit adjustment},
  booktitle    = {ICLR},
  year         = {2021},
}

@inproceedings{LA-02,
  author       = {Mengke Li and
                  Yiu{-}Ming Cheung and
                  Yang Lu},
  title        = {Long-tailed Visual Recognition via Gaussian Clouded Logit Adjustment},
  booktitle    = {CVPR},
  pages        = {6919--6928},
  year         = {2022},
}

@article{LA-03,
  title={Logit Normalization for Long-Tail Object Detection},
  author={ Zhao, Liang  and  Teng, Yao  and  Wang, Limin },
  journal={IJCV},
  volume={132},
  number={6},
  year={2024},
}

@inproceedings{LA-04,
  author       = {Xiaohang Xu and
                  Renhe Jiang and
                  Chuang Yang and
                  Zipei Fan and
                  Kaoru Sezaki},
  title        = {Taming the Long Tail in Human Mobility Prediction},
  booktitle    = {NeurIPS},
  year         = {2024},
}

@inproceedings{LA-05,
  author       = {Yuheng Jia and
                  Xiaorui Peng and
                  Ran Wang and
                  Min{-}Ling Zhang},
  title        = {Long-Tailed Partial Label Learning by Head Classifier and Tail Classifier Cooperation},
  booktitle    = {AAAI},
  pages        = {12857--12865},
  year         = {2024},
}

@inproceedings{LA-06,
  author       = {Wenjun Miao and
                  Guansong Pang and
                  Xiao Bai and
                  Tianqi Li and
                  Jin Zheng},
  title        = {Out-of-Distribution Detection in Long-Tailed Recognition with Calibrated Outlier Class Learning},
  booktitle    = {AAAI},
  pages        = {4216--4224},
  year         = {2024},
}

@article{Shannon-Entropy,
  author       = {Claude E. Shannon},
  title        = {A mathematical theory of communication},
  journal      = {Bell Syst. Tech. J.},
  volume       = {27},
  number       = {3},
  pages        = {379--423},
  year         = {1948},
}

@article{Tsallis-01,
  title={Possible generalization of Boltzmann-Gibbs statistics},
  author={Tsallis, Constantino},
  journal={Journal of statistical physics},
  volume={52},
  pages={479--487},
  year={1988},
}

@article{Tsallis-02,
  title={Generalized entropy-based criterion for consistent testing},
  author={Tsallis, Constantino},
  journal={Physical Review E},
  volume={58},
  number={2},
  pages={1442},
  year={1998},
}

@inproceedings{ImageNet-1k,   
    title={ImageNet: A large-scale hierarchical image database},  
    booktitle={CVPR},  
    author={Deng, Jia and Dong, Wei and Socher, Richard and Li, Li-Jia and  Kai Li and  Li Fei-Fei},  
    year={2009}, 
}

@inproceedings{ImageNet-V,  
 title={Do ImageNet Classifiers Generalize to ImageNet}, 
 booktitle={CVPR}, 
 author={Recht, Benjamin and Roelofs, Rebecca and Schmidt, Ludwig and Shankar, Venkatesh}, 
 year={2019}, 
}

@inproceedings{ImageNet-K,  
    title={Learning Robust Global Representations by Penalizing Local Predictive Power.}, 
    booktitle={CVPR}, 
    author={Wang, Haohan and Ge, Songwei and Xing, EricP. and Lipton, ZacharyC.}, 
    year={2019}, 
}

@inproceedings{ImageNet-A,  
    title={Natural Adversarial Examples}, 
    author={Hendrycks, Dan and Berkeley, U and Zhao, Kevin and Basart, Steven and Uchicago, Jacob and Steinhardt, Dawn and Song, U}, 
    year={2019}, 
    booktitle={arXiv}, 
}

@inproceedings{ImageNet-R,  
    title={The Many Faces of Robustness: A Critical Analysis of Out-of-Distribution Generalization}, 
    booktitle={arXiv}, 
    author={Hendrycks, Dan and Basart, Steven and Mu, Norman and Kadavath, Saurav and Wang, Frank and Dorundo, Evan and Desai, Rahul and Zhu, Tyler and Parajuli, Samyak and Guo, Mike and Song, Dezhen and Steinhardt, Jacob and Gilmer, Justin}, 
    year={2020}, 
}

@inproceedings{Caltech,  
title={Learning Generative Visual Models from Few Training Examples: An Incremental Bayesian Approach Tested on 101 Object Categories}, 
booktitle={CVPR}, 
author={ Li Fei-Fei and Fergus, R. and Perona, P.}, 
year={2005}, 
}

@inproceedings{SUN,  
title={SUN database: Large-scale scene recognition from abbey to zoo}, 
booktitle={CVPR}, 
author={Xiao, Jianxiong and Hays, James and Ehinger, Krista A. and Oliva, Aude and Torralba, Antonio}, 
year={2010}, 
}

@inproceedings{DTD,  
title={Describing Textures in the Wild.}, 
author={Cimpoi, Mircea and Maji, Subhransu and Kokkinos, Iasonas and Mohamed, Sammy and Vedaldi, Andrea}, 
year={2014},
booktitle={CVPR}, 
}

@article{EuroSAT,  
title={EuroSAT: A Novel Dataset and Deep Learning Benchmark for Land Use and Land Cover Classification}, 
journal={IEEE Journal of Selected Topics in Applied Earth Observations and Remote Sensing}, 
author={Helber, Patrick and Bischke, Benjamin and Dengel, Andreas and Borth, Damian}, 
year={2019}, 
pages={2217–2226}, 
}

@article{UCF,  
title={UCF101: A Dataset of 101 Human Actions Classes From Videos in The Wild}, 
journal={arXiv}, 
author={Soomro, Khurram and Zamir, Amir and Shah, Mubarak}, 
year={2012}, 
}

@inproceedings{Pets,  
title={Cats and dogs}, 
booktitle={CVPR}, 
author={Parkhi, O. M. and Vedaldi, A. and Zisserman, A. and Jawahar, C. V.}, 
year={2012}, 
}

@inproceedings{Cars,  
title={3D Object Representations for Fine-Grained Categorization}, 
booktitle={ICCV}, 
author={Krause, Jonathan and Stark, Michael and Deng, Jia and Fei-Fei, Li}, 
year={2013}, 
}

@inproceedings{Flowers,  
title={Automated Flower Classification over a Large Number of Classes}, 
booktitle={2008 Sixth Indian Conference on Computer Vision, Graphics and Image Processing}, 
author={Nilsback, Maria-Elena and Zisserman, Andrew}, 
year={2008}, 
}

@inproceedings{Food,  
title={Food-101 – Mining Discriminative Components with Random Forests}, 
booktitle={ECCV}, 
author={Bossard, Lukas and Guillaumin, Matthieu and Van Gool, Luc}, 
year={2014}, 
pages={446–461}, 
}

@article{Aircraft,  
title={Fine-Grained Visual Classification of Aircraft}, 
journal={arXiv}, 
author={Maji, Subhransu and Rahtu, Esa and Kannala, Juho and Blaschko, MatthewB. and Vedaldi, Andrea}, 
year={2013}, 
}

@inproceedings{CuPL,
  author       = {Sarah M. Pratt and
                  Ian Covert and
                  Rosanne Liu and
                  Ali Farhadi},
  title        = {What does a platypus look like? Generating customized prompts for
                  zero-shot image classification},
  booktitle    = {ICCV},
  pages        = {15645--15655},
  year         = {2023},
}

@inproceedings{bias-01,
  author       = {James Urquhart Allingham and
                  Jie Ren and
                  Michael W. Dusenberry and
                  Xiuye Gu and
                  Yin Cui and
                  Dustin Tran and
                  Jeremiah Zhe Liu and
                  Balaji Lakshminarayanan},
  title        = {A Simple Zero-shot Prompt Weighting Technique to Improve Prompt Ensembling
                  in Text-Image Models},
  booktitle    = {ICML},
  year         = {2023}
}

@article{bias-02,
  title={The Neglected Tails of Vision-Language Models},
  author={Parashar, Shubham and Lin, Zhiqiu and Liu, Tian and Dong, Xiangjue and Li, Yanan and Ramanan, Deva and Caverlee, James and Kong, Shu},
  journal={arXiv preprint arXiv:2401.12425},
  year={2024}
}

@inproceedings{bias-03,
  author       = {Beier Zhu and
                  Kaihua Tang and
                  Qianru Sun and
                  Hanwang Zhang},
  title        = {Generalized Logit Adjustment: Calibrating Fine-tuned Models by Removing
                  Label Bias in Foundation Models},
  booktitle    = {NeurIPS},
  year         = {2023}
}

@article{lee2024entropy,
  title={Entropy is not enough for test-time adaptation: From the perspective of disentangled factors},
  author={Lee, Jonghyun and Jung, Dahuin and Lee, Saehyung and Park, Junsung and Shin, Juhyeon and Hwang, Uiwon and Yoon, Sungroh},
  journal={arXiv preprint arXiv:2403.07366},
  year={2024}
}

@article{hendrycks2019benchmarking,
  title={Benchmarking neural network robustness to common corruptions and perturbations},
  author={Hendrycks, Dan and Dietterich, Thomas},
  journal={arXiv preprint arXiv:1903.12261},
  year={2019}
}

@article{lu2023meta,
  title={Meta-tsallis-entropy minimization: a new self-training approach for domain adaptation on text classification},
  author={Lu, Menglong and Huang, Zhen and Tian, Zhiliang and Zhao, Yunxiang and Fei, Xuanyu and Li, Dongsheng},
  journal={arXiv preprint arXiv:2308.02746},
  year={2023}
}

@article{liu2021cycle,
  title={Cycle self-training for domain adaptation},
  author={Liu, Hong and Wang, Jianmin and Long, Mingsheng},
  journal={Advances in Neural Information Processing Systems},
  volume={34},
  pages={22968--22981},
  year={2021}
}

@inproceedings{leeduet,
  title={DUET: Dual-Perspective Pseudo Labeling and Uncertainty-aware Exploration \& Exploitation Training for Source-Free Domain Adaptation},
  author={Lee, Jae Yun and Park, Jae Hyeon and Lee, Gyoomin and Kim, Bogyeong and Cha, Min Hee and Nam, Hyeok and Jeon, Joo Hyeon and Lee, Hyunse and Cho, Sung In},
  booktitle={The Thirty-ninth Annual Conference on Neural Information Processing Systems}
}

@article{zhao2023source,
  title={Source-free domain adaptation for privacy-preserving seizure prediction},
  author={Zhao, Yuchang and Feng, Shuai and Li, Chang and Song, Rencheng and Liang, Deng and Chen, Xun},
  journal={IEEE Transactions on Industrial Informatics},
  volume={20},
  number={2},
  pages={2787--2798},
  year={2023},
  publisher={IEEE}
}

@article{yu2025visual,
  title={Visual Multi-Agent System: Mitigating Hallucination Snowballing via Visual Flow},
  author={Yu, Xinlei and Xu, Chengming and Zhang, Guibin and He, Yongbo and Chen, Zhangquan and Xue, Zhucun and Zhang, Jiangning and Liao, Yue and Hu, Xiaobin and Jiang, Yu-Gang and others},
  journal={arXiv preprint arXiv:2509.21789},
  year={2025}
}

@inproceedings{zhao2024balf,
  title={Balf: Simple and efficient blur aware local feature detector},
  author={Zhao, Zhenjun},
  booktitle={Proceedings of the IEEE/CVF Winter Conference on Applications of Computer Vision},
  pages={3362--3372},
  year={2024}
}

@inproceedings{edstedt2024dedode,
  title={Dedode v2: Analyzing and improving the dedode keypoint detector},
  author={Edstedt, Johan and B{\"o}kman, Georg and Zhao, Zhenjun},
  booktitle={Proceedings of the IEEE/CVF Conference on Computer Vision and Pattern Recognition},
  pages={4245--4253},
  year={2024}
}

@article{zeng2025janusvln,
            title={JanusVLN: Decoupling Semantics and Spatiality with Dual Implicit Memory for Vision-Language Navigation},
            author={Zeng, Shuang and Qi, Dekang and Chang, Xinyuan and Xiong, Feng and Xie, Shichao and Wu, Xiaolong and Liang, Shiyi and Xu, Mu and Wei, Xing},
            journal={arXiv preprint arXiv:2509.22548},
            year={2025}
            }

@article{zeng2025FSDrive,
      title={FutureSightDrive: Thinking Visually with Spatio-Temporal CoT for Autonomous Driving},
      author={Shuang Zeng and Xinyuan Chang and Mengwei Xie and Xinran Liu and Yifan Bai and Zheng Pan and Mu Xu and Xing Wei},
      journal={arXiv preprint arXiv:2505.17685},
      year={2025}
      }

@article{ke2025early,
  title={Early warning of cryptocurrency reversal risks via multi-source data},
  author={Ke, Zong and Cao, Yuqing and Chen, Zhenrui and Yin, Yuchen and He, Shouchao and Cheng, Yu},
  journal={Finance Research Letters},
  pages={107890},
  year={2025},
  publisher={Elsevier}
}

@article{ouyang2024learn,
  title={Learn from global correlations: Enhancing evolutionary algorithm via spectral gnn},
  author={Ouyang, Kaichen and Ke, Zong and Fu, Shengwei and Liu, Lingjie and Zhao, Puning and Hu, Dayu},
  journal={arXiv preprint arXiv:2412.17629},
  year={2024}
}

@inproceedings{CMDGK,
  author       = {Shuang Wu and
                  Heng Liang and
                  Yong Zhang and
                  Yanlin Chen and
                  Ziyu Jia},
  title        = {A Cross-Modal Densely Guided Knowledge Distillation Based on Modality
                  Rebalancing Strategy for Enhanced Unimodal Emotion Recognition},
  booktitle    = {IJCAI},
  pages        = {4236--4244},
  year         = {2025},
}

@article{zhao2026non,
  title={Non-Intrusive Graph-Based Bot Detection for E-Commerce Using Inductive Graph Neural Networks},
  author={Zhao, Sichen and Xue, Zhiming and Qi, Yalun and Zeng, Xianling and Yu, Zihan},
  journal={arXiv preprint arXiv:2601.22579},
  year={2026}
}

@inproceedings{SePer,
  author       = {Lu Dai and
                  Yijie Xu and
                  Jinhui Ye and
                  Hao Liu and
                  Hui Xiong},
  title        = {SePer: Measure Retrieval Utility Through The Lens Of Semantic Perplexity
                  Reduction},
  booktitle    = {ICLR},
  year         = {2025},
}

@article{wu2025evaluation,
  title={Evaluation of tunnel rock mass integrity using multi-modal data and generative large model: Tunnel rip-gpt},
  author={Wu, Chen and Huang, Hongwei and Ni, Yi-Qing},
  journal={Available at SSRN 5348429},
  year={2025}
}

@inproceedings{Refining,
  title = 	 {{Refining Visual Perception for Decoration Display}: {A} Self-Enhanced Deep Captioning Model},
  author =       {Huang, Longfei and Wu, Xiangyu and Wang, Jingyuan and Guo, Weili and Yang, Yang},
  booktitle = 	 {ACML},
  pages = 	 {527--542},
  year = 	 {2024},
  volume = 	 {260},
  month = 	 {05--08 Dec},
}

@inproceedings{CoVLR,
  author={Wan, Fengqiang and Wu, Xiangyu and Guan, Zhihao and Yang, Yang},
  booktitle={ICME}, 
  title={CoVLR: Coordinating Cross-Modal Consistency and Intra-Modal Relations for Vision-Language Retrieval}, 
  year={2024},
  pages={1-6},
}

@inproceedings{Facilitating,
  author       = {Yang Yang and
                  Fengqiang Wan and
                  Qing{-}Yuan Jiang and
                  Yi Xu},
  title        = {Facilitating Multimodal Classification via Dynamically Learning Modality
                  Gap},
  booktitle    = {NeurIPS},
  year         = {2024},
}

@inproceedings{q2v,
  author={Wu, Xiangyu and Lu, Jianfeng and Li, Zhuanfeng and Xiong, Fengchao},
  booktitle={ICIP}, 
  title={Ques-to-Visual Guided Visual Question Answering}, 
  year={2022},
  pages={4193-4197},
}

@inproceedings{PVP,
  title={TAI++: Text as Image for Multi-Label Image Classification by Co-Learning Transferable Prompt},
  author={Wu, Xiangyu and Jiang, Qing-Yuan and Yang, Yang and Wu, Yi-Feng and Chen, Qing-Guo and Lu, Jianfeng},
  booktitle    = {IJCAI},
  year={2024}
}

@ inproceedings{RML,
  author       = {Qing{-}Yuan Jiang and
                  Longfei Huang and
                  Yang Yang},
  title        = {Rethinking Multimodal Learning from the Perspective of Mitigating
                  Classification Ability Disproportion},
  booktitle    = {NeurIPS},
  year         = {2025},
}

@inproceedings{RPRCR,
  author       = {Ruoxuan Li and
                  Xiangyu Wu and
                  Yang Yang},
  title        = {Noise Self-Correction via Relation Propagation for Robust Cross-Modal
                  Retrieval},
  booktitle    = {ACM MM},
  pages        = {4748--4757},
  year         = {2025},
}

@inproceedings{TaAM-CPT,
  author       = {Xiangyu Wu and
                  Feng Yu and
                  Yang Yang and
                  Jianfeng Lu},
  title        = {Text as Any-Modality for Zero-Shot Classification by Consistent Prompt
                  Tuning},
  booktitle    = {ACM MM},
  pages        = {8106--8115},
  year         = {2025},
}
